\documentclass{article}
\usepackage{graphicx} 
\usepackage{xcolor}
\usepackage{amsmath}
\usepackage{amsfonts}
\usepackage{amssymb}
\usepackage{booktabs}
\usepackage{algorithm}
\usepackage{subcaption}
\usepackage{algorithmic}
\usepackage{caption}
\usepackage{tcolorbox}
\usepackage{dblfloatfix}
\usepackage{placeins}
\tcbuselibrary{listings,skins}

\makeatletter
\newcommand\avsuminner[2]{%
  {\sbox0{$\m@th#1\sum$}%
   \vphantom{\usebox0}%
   \ooalign{%
     \hidewidth
     \smash{\vrule height\dimexpr\ht0+1pt\relax depth\dimexpr\dp0+1pt\relax}%
     \hidewidth\cr
     $\m@th#1\sum$\cr
   }%
  }%
}
\makeatother

\usepackage{iclr2026,times}
\title{LEVDA: Latent Ensemble Variational Data Assimilation via Differentiable Dynamics}

\iclrfinalcopy

\begin{document}

\maketitle

\begin{abstract}
Long-range geophysical forecasts are fundamentally limited by chaotic dynamics and numerical errors. While data assimilation can mitigate these issues, classical variational smoothers require computationally expensive tangent-linear and adjoint models. Conversely, recent efficient latent filtering methods often enforce weak trajectory-level constraints and assume fixed observation grids. To bridge this gap, we propose Latent Ensemble Variational Data Assimilation (LEVDA), an ensemble-space variational smoother that operates in the low-dimensional latent space of a pretrained differentiable neural dynamics surrogate. By performing four-dimensional ensemble-variational (4DEnVar) optimization within an ensemble subspace, LEVDA jointly assimilates states and unknown parameters without the need for adjoint code or auxiliary observation-to-latent encoders. Leveraging the fully differentiable, continuous-in-time-and-space nature of the surrogate, LEVDA naturally accommodates highly irregular sampling at arbitrary spatiotemporal locations. Across three challenging geophysical benchmarks, LEVDA matches or outperforms state-of-the-art latent filtering baselines under severe observational sparsity while providing more reliable uncertainty quantification. Simultaneously, it achieves substantially improved assimilation accuracy and computational efficiency compared to full-state 4DEnVar.
\end{abstract}

\section{Introduction}
The accuracy of long-range forecasts in many geophysical systems is fundamentally limited by intrinsic chaos and a sensitive dependence on initial conditions \citep{lorenz2017deterministic, lorenz1969atmospheric, lorenz1982atmospheric}. Global weather prediction is a prominent example, but similar challenges arise in ocean dynamics, geophysical flows, and other high-dimensional spatiotemporal processes. In both traditional numerical models and newer machine-learning surrogates, forecast skill is further degraded by numerical errors stemming from uncertainty and discretization \citep{bauer2015quiet, dueben2018challenges}. Data assimilation (DA) mitigates these issues by systematically incorporating observational data into predictive models to infer initial conditions---and potentially unknown parameters---that are consistent with historical measurements, thereby reducing error growth over longer lead times \citep{kalnay2003atmospheric, carrassi2018data, bannister2017review}.

Recent score-based approaches, such as Latent-EnSF \citep{si2025latent} and LD-EnSF \citep{xiao2026ld}, have demonstrated strong performance for filtering under sparse observations by leveraging learned generative priors in a low-dimensional latent space. However, because these methods typically operate sequentially, they provide weaker trajectory-level constraints over a multi-time assimilation window—a critical drawback when enforcing dynamical consistency or performing reliable joint state--parameter estimation. Furthermore, they often require the training of additional amortized components, such as observation-to-latent encoders, whose underlying assumptions can be restrictive. While score-based smoothing methods like SDA \citep{rozet2023score} and FlowDAS \citep{chen2025flowdas} target the smoothing posterior directly, they can be computationally intensive and frequently treat dynamics implicitly rather than as an explicit, differentiable prior. Similarly, approaches like APPA \citep{andry2025appa} that learn latent dynamics via spatiotemporal diffusion models remain challenging to adapt for joint state--parameter estimation and highly irregular spatiotemporal observation networks.

Conversely, variational data assimilation methods, such as 4DVar \citep{rabier1998extended}, enforce strict dynamical consistency by fitting trajectories to all observations within a specified time window. Their primary computational bottleneck is the need for repeated nonlinear model runs and the derivation of gradients via explicit tangent-linear and adjoint models. Ensemble formulations like four-dimensional ensemble-variational (4DEnVar) method \citep{zhu2022four} mitigate this by replacing adjoints with flow-dependent covariance information derived from an ensemble. While recent machine-learning--accelerated variational methods (e.g., FengWu-4DVar \citep{xiao2024fengwu}) substantially speed up assimilation by reducing forward and gradient costs, they typically yield a single deterministic MAP trajectory; extracting calibrated uncertainty quantification (UQ) requires additional approximations \citep{hodyss2016extent}. Thus, a critical gap remains: achieving the rigorous multi-time constraints of variational smoothing and the reliable UQ of ensemble methods, while maintaining the computational efficiency of low-dimensional latent surrogates.

To bridge this gap, we introduce \emph{Latent Ensemble Variational Data Assimilation} (LEVDA), a smoothing-based framework that performs 4DEnVar-style ensemble-space optimization directly in the latent space of a pretrained differentiable dynamical surrogate (e.g., an LDNet \citep{regazzoni2024learning}). LEVDA enables gradient-based assimilation through the surrogate dynamics without requiring adjoint models, and it avoids the need to train auxiliary observation-to-latent encoders by computing observation misfits directly from decoded latent trajectories. By leveraging the fully differentiable and continuous-in-time-and-space nature of the surrogate, LEVDA effortlessly accommodates observations at arbitrary, off-grid spatiotemporal locations and natively supports the joint assimilation of states and unknown parameters. Consequently, LEVDA confers several practical benefits that are difficult to achieve simultaneously with prior data-driven approaches: it enforces multi-time dynamical constraints, handles highly irregular observation operators, and substantially reduces computational complexity compared to full-state ensemble variational smoothers.

In summary, our main contributions are:
\begin{itemize}
    \item We propose LEVDA, a latent extension of 4DEnVar that performs efficient variational smoothing in an ensemble subspace of a pretrained differentiable latent dynamics network.
    \item We enable joint state--parameter ensemble variational data assimilation without requiring explicit adjoint code or auxiliary observation-to-latent encoders.
    \item We show that LEVDA's continuous-in-time-and-space formulation naturally accommodates highly irregular, time-varying observations at arbitrary spatiotemporal locations.
    \item We evaluate LEVDA on three challenging geophysical benchmarks, demonstrating state-of-the-art accuracy under extreme observational sparsity with improved uncertainty quantification, while substantially reducing runtime costs relative to full-state 4DEnVar.
\end{itemize}

\section{Background}

We briefly review the modeling assumptions and variational frameworks that form the foundation of LEVDA. We first define the standard state-space model and observation process, followed by a summary of variational smoothing (4DVar) and its ensemble-based counterpart (4DEnVar). Finally, we introduce the differentiable latent dynamical surrogates that make efficient, low-dimensional assimilation possible.

\subsection{Dynamical Systems}

Consider a physical system characterized by a state vector $x_t \in \mathbb{R}^{d_x}$ and potentially time-varying parameters $u_t \in \mathbb{R}^{d_u}$ at discrete time steps $t \in \mathbb{Z}^+$. Given initial conditions $(x_0, u_0)$, we model the forward dynamics as a Markov process:
\begin{align}
    x_t &= M(x_{t-1}, u_{t-1}) + \epsilon_t, \quad \epsilon_t \sim \mathcal{N}(0, Q_t), \label{eq:state_dyn} \\
    u_t &= M_u(u_{t-1}) + \eta_t, \quad \eta_t \sim \mathcal{N}(0, Q_t^u), \label{eq:param_dyn}
\end{align}
for $t = 1, 2, \dots$, where the state transition $M: \mathbb{R}^{d_x} \times \mathbb{R}^{d_u} \to \mathbb{R}^{d_x}$ and parameter evolution $M_u: \mathbb{R}^{d_u} \to \mathbb{R}^{d_u}$ may be nonlinear. For strictly static parameters, we simply take $M_u(u) = u$ and $Q_t^u = 0$. 

Observations of the system are acquired via
\begin{equation}
    y_t = H_t(x_t) + \gamma_t, \qquad \gamma_t \sim \mathcal{N}(0, R_t), \label{eq:observation}
\end{equation}
where the observation operator $H_t: \mathbb{R}^{d_x} \to \mathbb{R}^{d_y}$ can be time-varying to accommodate irregular sampling or moving sensors. We assume the noise sequences $\{\epsilon_t\}$, $\{\eta_t\}$, and $\{\gamma_t\}$ are mutually independent, with covariance matrices that may vary over time.

\subsection{4DVar}

To unify the notation for joint state-parameter estimation, we define the augmented state vector $z_t = (x_t, u_t) \in \mathbb{R}^{d_x + d_u}$. Under the dynamical model in Equations \eqref{eq:state_dyn} and \eqref{eq:param_dyn}, the one-step transition densities and observation likelihood are standard Gaussians:
\begin{equation}
\begin{aligned}
    p(x_t \mid x_{t-1}, u_{t-1}) &= \mathcal{N}\bigl(M(x_{t-1}, u_{t-1}), Q_t\bigr), \\
    p(u_t \mid u_{t-1}) &= \mathcal{N}\bigl(M_u(u_{t-1}), Q_t^u\bigr), \\
    p(y_t \mid x_t) &= \mathcal{N}\bigl(H_t(x_t), R_t\bigr).
\end{aligned}
\end{equation}
Equivalently, the augmented transition density factors as $p(z_t \mid z_{t-1}) = p(x_t \mid x_{t-1}, u_{t-1}) p(u_t \mid u_{t-1})$, and the likelihood depends only on the physical state, $p(y_t \mid z_t) = p(y_t \mid x_t)$. 

Given a sequence of past observations $y_{1:t-1}$ and a fixed assimilation window of length $\tau$, the smoothing posterior over the window is given by Bayes' rule:
\begin{equation}
    p(z_{t:t+\tau} \mid y_{1:t+\tau}) \propto p(z_{t:t+\tau} \mid y_{1:t-1}) \, p(y_{t:t+\tau} \mid z_{t:t+\tau}).
    \label{eq:smoothing_posterior_bayes}
\end{equation}
Because $y_{t:t+\tau}$ is conditionally independent of $y_{1:t-1}$ given the trajectory $z_{t:t+\tau}$, the likelihood term simplifies. Applying the Markov property of $\{z_t\}$ and the conditional independence of the observations, the posterior neatly factorizes as:
\begin{equation}
    p(z_{t:t+\tau} \mid y_{1:t+\tau}) \propto p(z_t \mid y_{1:t-1}) \prod_{i=1}^\tau p(z_{t+i} \mid z_{t+i-1}) \prod_{i=0}^\tau p(y_{t+i} \mid x_{t+i}).
    \label{eq:smoothing_posterior_factorized}
\end{equation}

Variational smoothing, or 4DVar, computes a maximum a posteriori (MAP) estimate of the trajectory over the assimilation window. In incremental implementations, the filtering prior is approximated by a Gaussian background, $p(z_t \mid y_{1:t-1}) \approx \mathcal{N}(z_t^b, B)$, and the optimal state is found by minimizing the negative log-posterior. Under the Gaussian assumptions established above, the weak-constraint 4DVar cost function (up to an additive constant) is given by:
\begin{equation}
\begin{aligned}
    J(z_{t:t+\tau}) = & \; \frac{1}{2} \| z_t - z_t^b \|_{B^{-1}}^2 + \frac{1}{2} \sum_{i=1}^{\tau} \| x_{t+i} - M(x_{t+i-1}, u_{t+i-1}) \|_{Q_{t+i}^{-1}}^2 \\
    & + \frac{1}{2} \sum_{i=1}^{\tau} \| u_{t+i} - M_u(u_{t+i-1}) \|_{(Q_{t+i}^u)^{-1}}^2  + \frac{1}{2} \sum_{i=0}^{\tau} \| y_{t+i} - H_{t+i}(x_{t+i}) \|_{R_{t+i}^{-1}}^2.
\end{aligned}
\end{equation}
Strong-constraint 4DVar is recovered by assuming a perfect model ($Q_t = 0$ and $Q_t^u = 0$). In this setting, the future states $(x_{t+i}, u_{t+i})$ are strictly determined by the initial condition $(x_t, u_t)$ via the forward dynamics, allowing the optimization to be posed directly over $z_t$.

\subsection{4DEnVar}

4DEnVar is an ensemble-space analogue of incremental 4DVar that avoids the need for explicit tangent-linear and adjoint models by restricting the analysis increment to a subspace spanned by an ensemble \citep{courtier1994strategy, zhu2022four}. Maintaining consistency with the previous subsection, we work with the augmented state $z_t = (x_t, u_t)$. In strong-constraint 4DEnVar, the model is assumed perfect over the window, meaning the full trajectory is completely determined by the window-initial condition.

Let $z_t^b$ be a background initial condition (e.g., the mean of the filtering prior) and let $\{z_t^{(k)}\}_{k=1}^K$ be an ensemble of initial conditions with mean $\bar{z}_t = \frac{1}{K}\sum_{k=1}^K z_t^{(k)}$. We define the perturbations as $\delta z_t^{(k)} = z_t^{(k)} - \bar{z}_t$, which form the ensemble perturbation matrix:
\begin{equation} 
P_z = \frac{1}{\sqrt{K-1}} \begin{bmatrix} 
\delta z_t^{(1)} & \cdots & \delta z_t^{(K)} 
\end{bmatrix}.
\end{equation}
This formulation provides an empirical approximation of the background covariance, $B \approx P_z P_z^T$. Consequently, 4DEnVar parameterizes the analysis increment at the window start as $\delta z_t = P_z \alpha$ for a coefficient vector $\alpha \in \mathbb{R}^K$.

To evaluate the observation-space misfit, each ensemble member is propagated through the nonlinear model to obtain trajectories $\{z_{t+i}^{(k)}\}_{k=1}^K$ for $i=0,\dots,\tau$. Let $x_{t+i}^{(k)}$ denote the physical state component of $z_{t+i}^{(k)}$. For each observation time, we compute the observation-space perturbations:
\begin{equation}
\delta y_{t+i}^{(k)} = H_{t+i}(x_{t+i}^{(k)}) - \frac{1}{K}\sum_{\ell=1}^K H_{t+i}(x_{t+i}^{(\ell)}), \quad i=0,\dots,\tau,
\end{equation}    
and assemble the corresponding matrices \begin{equation}
P_{y,t+i} = \frac{1}{\sqrt{K-1}} [ \delta y_{t+i}^{(1)} \cdots \delta y_{t+i}^{(K)} ].
\end{equation}
Let $x_{t+i}^b$ be the background trajectory obtained by propagating $z_t^b$, and define the innovations $d_{t+i} = y_{t+i} - H_{t+i}(x_{t+i}^b)$. The incremental 4DEnVar objective in the reduced ensemble space is then:
\begin{equation}
J(\alpha) = \frac{1}{2} \alpha^T \alpha + \frac{1}{2} \sum_{i=0}^{\tau} \| d_{t+i} - P_{y,t+i} \alpha \|_{R_{t+i}^{-1}}^2.
\end{equation}
Minimizing this quadratic cost yields the optimal coefficients $\alpha^*$, and the analysis update is computed as $z_t^a = z_t^b + P_z \alpha^*$ (yielding the updated physical states and parameters $(x_t^a, u_t^a)$). Operationally, outer loops are often used to update the background trajectory and recompute perturbations, addressing nonlinearities in the dynamics and observation operators. Additionally, localization and inflation are typically applied to maintain filter stability in high-dimensional regimes. Weak-constraint variants can be constructed by augmenting the ensemble-space control with model-error terms, analogous to the 4DVar formulation above.

\subsection{Differentiable Latent Dynamics}
Latent Dynamics Networks (LDNets) \citep{regazzoni2024learning} provide a compact, differentiable surrogate for high-dimensional dynamical systems by jointly learning a latent vector field and a decoder to physical space. We represent the latent state as a continuous-time function $s(t) \in \mathbb{R}^{d_s}$ and let $u(t)$ denote any (known or unknown) time-varying parameters or forcings. A typical LDNet defines the latent evolution via an ordinary differential equation:
\begin{equation}
    \frac{d}{dt} s(t) \;=\; \mathcal{F}_\theta(s(t), u(t)),
    \label{eq:dyn_net}
\end{equation}
where the neural vector field $\mathcal{F}_\theta$ can be integrated using a differentiable numerical ODE solver. For example, discretizing time on a regular grid $t_k = t_0 + k \Delta t$ and assuming $u(t)$ is piecewise constant yields the forward-Euler update:
\begin{equation}
    s_{k+1} \;=\; s_k + \Delta t \, \mathcal{F}_\theta(s_k, u_k), \qquad k=0,1,\dots,
    \label{eq:latent_euler}
\end{equation}
where $s_k \approx s(t_k)$ and $u_k \approx u(t_k)$.

Given a latent trajectory $s(t)$, a decoder $D_\phi$ maps the latent states back to physical quantities. In particular, we use coordinate-based neural field decoders that enable continuous spatial evaluation at any query point $\xi \in \Omega$ via
\begin{equation}
\hat{x}(t, \xi) = D_\phi(s(t), \xi).    
\end{equation} 
Following \citet{xiao2026ld}, we adopt a shifted discrete-time convention with a fixed initialization $s_{-1} = 0$ to naturally accommodate varying initial conditions across different trajectories. Furthermore, the integration step size $\Delta t$ can be treated as a learnable parameter and optimized jointly with the network weights $(\theta, \phi)$.

To train the surrogate, we assume access to a dataset of $N$ physical trajectories $\{x^{(n)}(t_k, \cdot)\}_{k=0}^T$ with potentially varying initial conditions. The model is trained to minimize the reconstruction error of the decoded latent trajectories. To reduce computational cost and naturally handle irregular spatial data, we sample a subset of spatial query points $\Xi_{n,k} \subset \Omega$ for each trajectory $n$ at time $t_k$, evaluating the decoder only at these locations. Denoting the predicted field as $\hat{x}^{(n)}(t_k, \xi) = D_\phi(s^{(n)}(t_k), \xi)$, the standard training objective is:
\begin{equation}
    \mathcal{L}(\theta, \phi) = \frac{1}{N} \sum_{n=1}^N \sum_{k=0}^T \frac{1}{|\Xi_{n,k}|} \sum_{\xi \in \Xi_{n,k}} \left\| \hat{x}^{(n)}(t_k, \xi) - x^{(n)}(t_k, \xi) \right\|_2^2.
    \label{eq:latent_loss}
\end{equation}
Here, each latent trajectory $s^{(n)}(t_k)$ is generated by integrating the learned dynamics forward. Crucially, this combination of continuous-time latent evolution, coordinate-based decoding, and end-to-end differentiability forms the technical foundation of LEVDA, enabling gradient-based variational smoothing across arbitrary, grid-free spatiotemporal observation networks.

\section{Methodology}

LEVDA integrates differentiable latent dynamics with ensemble-variational smoothing to formulate a latent-space analogue of 4DEnVar. For a given assimilation window starting at time $t$, the method takes three inputs: (i) a fully differentiable pretrained latent surrogate (trained via \eqref{eq:latent_loss}), (ii) an ensemble representing the background distribution of window-initial latent variables, and (iii) a collection of (potentially spatiotemporally irregular) observations. In turn, LEVDA produces an analysis ensemble of latent states and parameters for the window onset, along with decoded physical trajectories that harmonize the observations with the latent dynamics.

\subsection{Latent-4DVar}\label{sec:latent-4dvar}

We assume access to a pretrained differentiable latent surrogate that evolves an augmented latent state $z(t) = (s(t), u(t))$ and decodes $s(t)$ to physical space via $D_\phi$. Concretely, we model the latent dynamics by
\begin{equation}
    \frac{d}{dt} s(t) = \mathcal{F}_\theta(s(t), u(t)), \qquad \frac{d}{dt} u(t) = \mathcal{G}(u(t)), 
    \label{eq:latent-surrogate}
\end{equation}
where $\mathcal{G}$ encodes any assumed parameter evolution (e.g., $\mathcal{G} \equiv 0$ for static parameters). The decoded physical state is defined as $x(t) = D_\phi(s(t))$. Observations at time $t_i$ follow the standard state-space model from Equation \eqref{eq:observation}; crucially, however, the observation operator $H_{t_i}$ is applied in physical space \emph{after} decoding. This means LEVDA does not require an additional observation-to-latent encoder as used in LD-EnSF \citep{xiao2026ld}. The assimilation process simply differentiates through the composite map $z(t) \mapsto s(t) \mapsto x(t) = D_\phi(s(t)) \mapsto H_{t_i}(x(t_i))$.

For a smoothing window anchored at time $t$, let $\mathcal{T}_t = \{{t}_1, \dots, {t}_m\}$ denote the sequence of observation times within the window. These observations can be irregular and need not lie on a fixed grid. To handle arbitrary observation times, any $\tilde{t} \in \mathcal{T}_t$ falling between the fixed discrete integration steps $t_k$ and $t_{k+1}$ of \eqref{eq:latent_euler} is parameterized as $\tilde{t} = t_k + \alpha \Delta t$ with $\alpha \in [0, 1)$. Consistent with a forward-Euler discretization, we approximate the latent state at the observation time via linear interpolation:
\begin{equation}
    s(\tilde{t}) \approx s(t_k) + \alpha \Delta t \dot{s}(t_k), \label{eq:interp}
\end{equation}
where $\dot{s}(t_k) = \mathcal{F}_\theta(s(t_k), u(t_k))$. Because this continuous-time formulation gracefully handles off-grid data, LEVDA naturally accommodates a varying number of available observations per assimilation window.

Latent-space 4DVar computes a maximum a posteriori (MAP) estimate of the window-initial latent variables $z(t) = (s(t), u(t))$, which in turn determines the implied trajectory under Equation \eqref{eq:latent-surrogate}. We approximate the filtering prior with a Gaussian background $z(t) \sim \mathcal{N}(z_t^b, B)$, where $z_t^b = (s_t^b, u_t^b)$, and assume a block-diagonal covariance $B = \mathrm{diag}(B_s, B_u)$. In practice, $B_s$ and $B_u$ can be estimated from the background ensemble (e.g., using a diagonal or shrinkage estimator). The strong-constraint latent 4DVar objective is therefore:
\begin{equation}
    J(s(t), u(t)) = \frac{1}{2} \| s(t) - s_t^b \|^2_{B_s^{-1}} + \frac{1}{2} \| u(t) - u_t^b \|^2_{B_u^{-1}} + \frac{1}{2} \sum_{t_i \in \mathcal{T}_t} \bigl\| y_{t_i} - H_{t_i}\bigl(D_\phi(s(t_i))\bigr) \bigr\|^2_{R_{t_i}^{-1}},
    \label{eq:latent-4dvar-obj}
\end{equation}
where the state $s(t_i)$ at each observation time is obtained by integrating Equation \eqref{eq:latent-surrogate} forward from the initial condition $(s(t), u(t))$. While one could optimize over the full latent trajectory and include a weak-constraint model-mismatch term (analogous to weak-constraint 4DVar), we focus here on the strong-constraint setting enabled by the differentiable surrogate.

\subsection{Latent Ensemble Variational Data Assimilation}

While latent-space 4DVar provides a clean MAP formulation, its direct application requires specifying (and inverting) background covariances over the augmented latent variables $(s,u)$ and optimizing in the full augmented latent dimension $d_s+d_u$. LEVDA is designed to make latent variational smoothing computationally efficient and robust by: (i) restricting the update to an ensemble subspace, thereby optimizing over only $K$ coefficients; (ii) avoiding the explicit construction or inversion of $B_s$ and $B_u$ by relying on the empirical covariance of the ensemble; (iii) leveraging a flow-dependent, data-adaptive geometry via ensemble perturbations; and (iv) returning an analysis \emph{ensemble} for uncertainty quantification, rather than a single deterministic MAP trajectory.

Let $\{z_t^{b,(k)}\}_{k=1}^K$ be an ensemble of background window-initial latent variables, where $z_t^{b,(k)} = (s_t^{b,(k)}, u_t^{b,(k)})$ and the ensemble mean is $\bar{z}_t^b = \frac{1}{K}\sum_{k=1}^K z_t^{b,(k)}$. Operationally, this background ensemble can be obtained by propagating the previous analysis ensemble forward under the latent surrogate, or by encoding an ensemble of full-state estimates into the latent space (if an encoder is available). We define the perturbations as $\delta z_t^{(k)} = z_t^{b,(k)} - \bar{z}_t^b$. To prevent filter collapse, we optionally apply multiplicative and additive inflation:
\[
    \delta z_t^{(k)} \leftarrow \rho \, \delta z_t^{(k)} + \epsilon^{(k)}, \quad \rho \ge 1, \ \ \epsilon^{(k)} \sim \mathcal{N}(0, \sigma^2 I_{d_s+d_u}),
\]
and then form the ensemble perturbation matrix:
\[
    P_{z} \;=\; \frac{1}{\sqrt{K-1}} \begin{bmatrix} \delta z_t^{(1)} & \cdots & \delta z_t^{(K)} \end{bmatrix}.
\]

LEVDA restricts candidate analysis initial conditions to the affine subspace $z_t(\alpha) = \bar{z}_t^b + P_{z} \alpha$ for $\alpha \in \mathbb{R}^K$. This restriction implicitly assumes a standard Gaussian prior $\alpha \sim \mathcal{N}(0,I)$ and an empirical background covariance $B \approx P_{z} P_{z}^T$. Given a candidate state $z_t(\alpha)$, we integrate Equation \eqref{eq:latent-surrogate} to obtain the latent trajectory and its decoded states. We then minimize the following ensemble-space variational objective:
\begin{equation}
    \begin{aligned}
        J(\alpha) =& \; \frac{1}{2} \|\alpha\|_2^2 + \frac{1}{2} \sum_{t_i \in \mathcal{T}_t} \big\|y_{t_i} - H_{t_i}\!\bigl(D_{\phi}(s(t_i; \alpha))\bigr)\big\|_{R_{t_i}^{-1}}^2,
    \end{aligned}
    \label{eq:levda-obj}
\end{equation}
where $s(t_i; \alpha)$ denotes the latent state at time $t_i$ obtained by integrating forward from the initial condition $z_t(\alpha)$.

To produce an analysis ensemble rather than a single point estimate, we perform multiple independent optimizations of Equation \eqref{eq:levda-obj} using different warm starts within the same ensemble subspace. Specifically, we optimize a collection of coefficient vectors $\{\alpha^{(j)}\}_{j=1}^K$. Each vector is initialized to represent one of the background members (e.g., starting at $\alpha^{(j)} = \sqrt{K-1}e_j$, so that $\bar{z}_t^b + P_{z} \alpha^{(j)}$ recovers the $j$-th background member when $\epsilon^{(k)}=0$). We solve each optimization problem using gradient-based methods (e.g., L-BFGS) via automatic differentiation through the latent solver. Because the observation misfit remains nonlinearly coupled to $\alpha$, these distinct optimization trajectories allow the analysis members to capture local nonlinearity and multi-modality. Furthermore, this process is trivially parallelizable across the $K$ members. Finally, we construct the analysis ensemble by setting 
$$
z_t^{a,(j)} = \bar{z}_t^b + P_{z} \alpha_*^{(j)}, \quad j=1,\dots,K,
$$ 
yielding the updated parameters $u_t^{a,(j)}$ and decoded analysis states $x_t^{a,(j)} = D_\phi(s_t^{a,(j)})$.

\textbf{Connection to 4DEnVar.} LEVDA shares the same ensemble-subspace philosophy as 4DEnVar: both restrict the analysis increment to an affine subspace spanned by background ensemble perturbations. The key distinction lies in how sensitivities are handled. Classical 4DEnVar avoids explicit tangent-linear and adjoint models by using nonlinear ensemble propagation to form an empirical linear map from ensemble coefficients to observation-space increments (e.g., via $P_{y,t_i}$), which yields a purely quadratic ensemble-space objective that can be solved without differentiating through the forward model. In contrast, LEVDA keeps the observation misfit nonlinearly coupled to the coefficients through the differentiable latent dynamics and decoder. It computes $\nabla_\alpha J$ in Equation \eqref{eq:levda-obj} by exact automatic differentiation through the latent solver, $D_\phi$, and $H_{t_i}$. Thus, LEVDA avoids the need for adjoint code for the original high-dimensional simulator, but fully leverages the exact gradients of the pretrained differentiable surrogate. When the latent dimension approaches the full state dimension and $D_\phi$ is close to the identity map, LEVDA closely mirrors a full-state ensemble-variational smoother. In the low-dimensional setting, however, it yields substantial computational savings while retaining rigorous, multi-time, dynamics-aware constraints.

\section{Experiments}

\subsection{Experimental Setup}

We evaluate LEVDA on three benchmark settings adapted from \citet{xiao2026ld}: (i) Kolmogorov flow with uncertain viscosity, (ii) tsunami propagation with an unknown source, and (iii) global atmospheric dynamics with an unknown forcing term. 

\textbf{Kolmogorov flow.} Canonical sinusoidally forced turbulence with an unknown viscosity (or Reynolds number $Re$)~\citep{kochkov2021machine}. The evaluation window covers 200 simulation steps following a 100-step burn-in period.

\textbf{Tsunami modeling.} Simplified shallow-water dynamics where an initial Gaussian bump is placed at a random location to simulate an unknown landslide or earthquake source~\citep{si2025latent}.

\textbf{Atmospheric modeling.} A global spherical simulation adapted from the \texttt{planetswe} example in \texttt{Well}~\citep{ohana2024well}, governed by complex dynamics with an unknown forcing term. The simulation spans 21 days ($504$ hours), initialized from ERA5 500\,hPa data~\citep{hersbach2020era5}. Observations are extremely sparse, covering only $\sim 0.2\%$ of the time steps.

Table~\ref{tab:exp-settings} summarizes the discretization details and spatial and temporal observation sparsity for each.
\begin{table*}[h]
\centering
\footnotesize
\setlength{\tabcolsep}{6pt}
\renewcommand{\arraystretch}{1.2}
\begin{tabular}{@{} l c c c c c}
\toprule
\textbf{Setting} & \textbf{Grid} & \textbf{Simulation steps} & \textbf{Obs. times (interval)} & $\Delta t$ & \textbf{Obs. grid}\\
\midrule
Kolmogorov flow & $150\times150$ & 200 & 40 (5) & 0.04 & $10\times10$\\
Tsunami modeling & $150\times150$ & 2000  & 50 (40) & $\approx 21$ s & $10\times10$\\
Atmospheric modeling & $512\times256$ & 30240 & 63 (480) & $\approx 60$ s & $16\times8$\\
\bottomrule
\end{tabular}
\caption{Summary of experimental setups (adapted from \citet{xiao2026ld}). {Obs. times} represent the total number of observation times while interval in parentheses denotes the spacing (in simulation steps) between consecutive observations. {Obs. grid} indicates the spatial subsampling resolution.}
\label{tab:exp-settings}
\end{table*}

\textbf{Implementation Details.}
For all experiments, we use the LDNets from \citet{xiao2026ld} with latent dimensions $D_s=10$ (Kolmogorov), $D_s=12$ (Tsunami), and $D_s=52$ (Atmosphere).
Unless otherwise stated, we use an ensemble size of $K=100$ for all latent methods (LEVDA, LD-EnSF, Latent-EnSF) and $K=40$ for the full-state 4DEnVar baseline (constrained by high cost of full-state propagation).
Observations are generated by subsampling the ground truth (see Table~\ref{tab:exp-settings}) and adding Gaussian noise with a $10\%$ noise-to-signal ratio.

\textbf{Metric.}
We evaluate reconstruction accuracy using the relative Root Mean Square Error (RMSE) of the analysis ensemble mean $\bar{x}_t$ with respect to the ground truth $x_t^{\star}$ measured in the Euclidean norm:
\[
  \mathrm{relative}\; \mathrm{RMSE}(t) \;=\; \frac{\|\bar{x}_t - x_t^{\star}\|_2}{\|x_t^{\star}\|_2}.
\]
In time-series plots, shaded regions denote the standard deviation of the member-wise relative errors, which are
\[
e^{(i)}_t \;=\; \frac{\|x^{(i)}_t - x_t^{\star}\|_2}{\|x_t^{\star}\|_2}, \qquad i=1,\dots,N.
\]
\subsection{Baselines}

We compare LEVDA against five baselines spanning state-space filtering, latent-space filtering, and full-state smoothing. All methods use the same observations and noise level.

\textbf{State-Space Filtering.} We use local ensemble transform Kalman filter (LETKF) \citep{hunt2007efficient} and ensemble score filter (EnSF) \citep{bao2024ensemble}. These operate directly on the high-dimensional grid but often struggle with sparse observations and nonlinear dynamics.

\textbf{Latent-Space Filtering.} We compare against {Latent-EnSF}~\citep{si2025latent}, which filters in a latent space (requiring coupled state and observation encoders), and {LD-EnSF}~\citep{xiao2026ld}, which uses the same differentiable dynamics as LEVDA but in a sequential filtering framework relying on an LSTM observation encoder.

\textbf{Full-State Smoothing.} We use {4DEnVar} \citep{zhu2022four} as a high-dimensional gold standard. We report results only for the tsunami setting because of the substantial engineering and runtime costs required at our target resolutions. Specifically, full-state 4DEnVar operates directly in the state space, requiring the construction of localized covariance operators and careful tuning of localization and inflation parameters. These steps are tightly coupled to the simulator’s discretization and boundary conditions, introducing significant implementation complexity and overhead.

\subsection{Comparison of Accuracy, Robustness, and Uncertainty}

\begin{figure*}[h]
	    \centering
	    \includegraphics[width=0.33\linewidth]{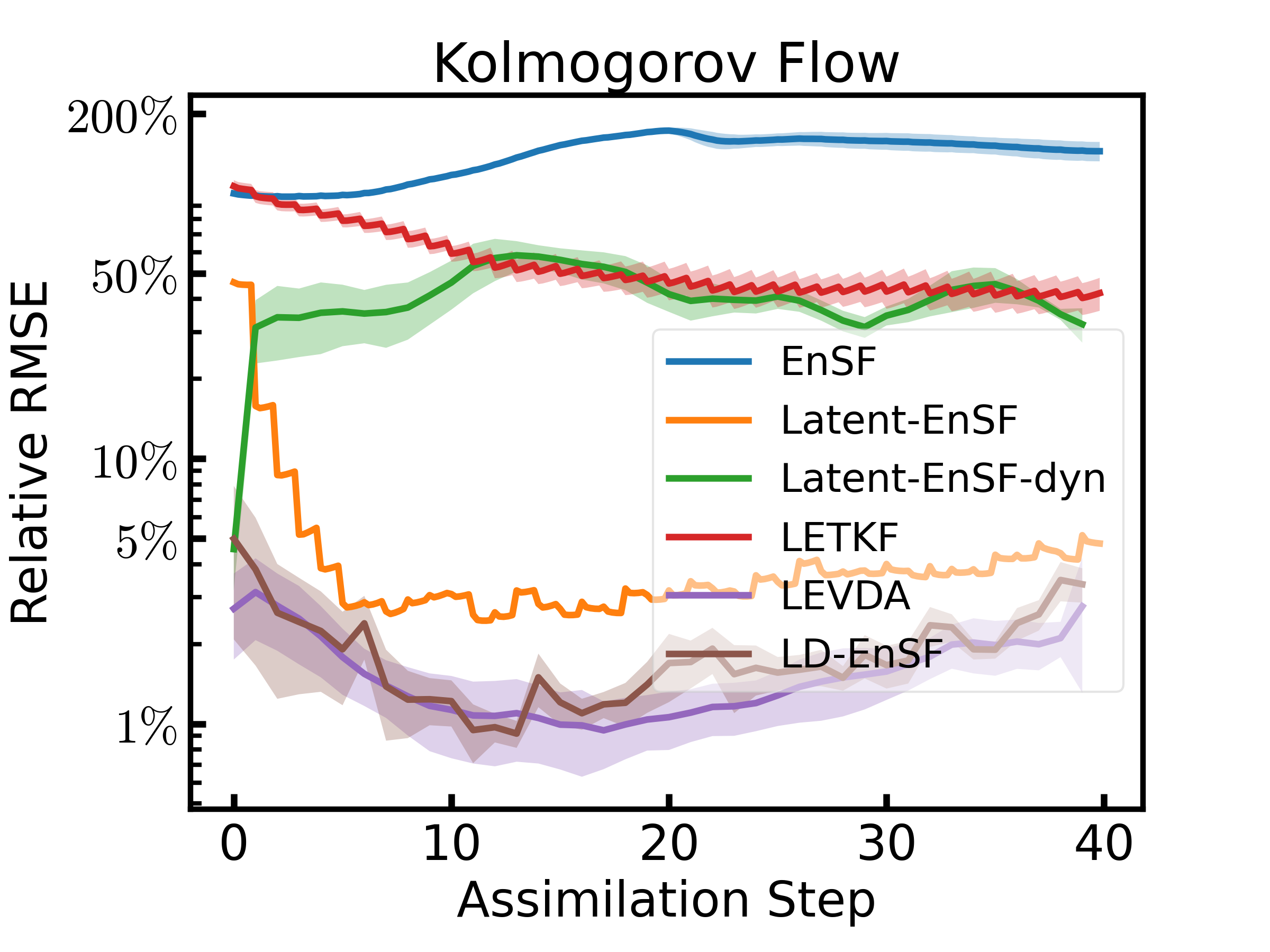}\hspace{-0.2em}
    \includegraphics[width=0.33\linewidth]{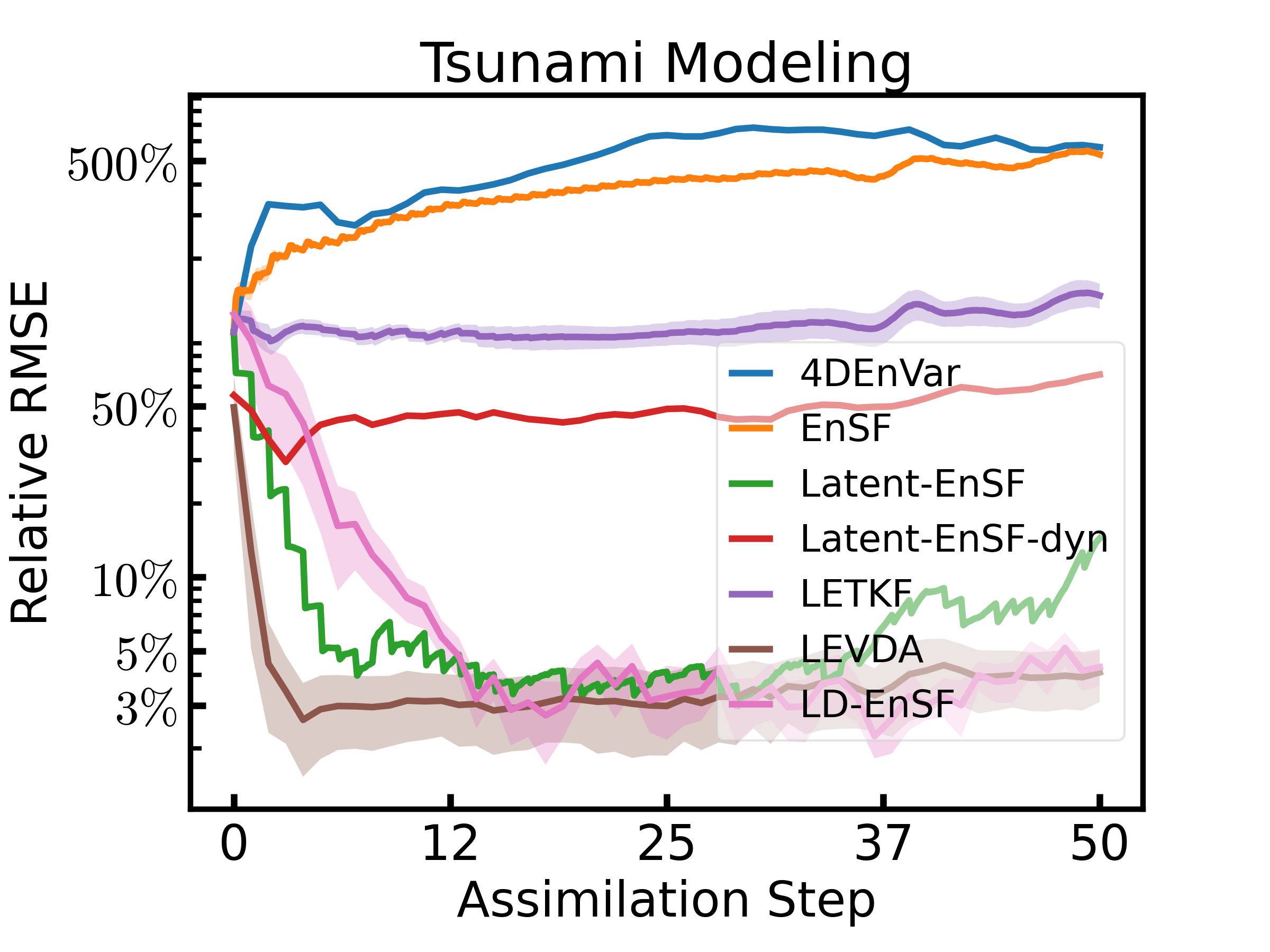}\hspace{-0.2em}
    \includegraphics[width=0.33\linewidth]{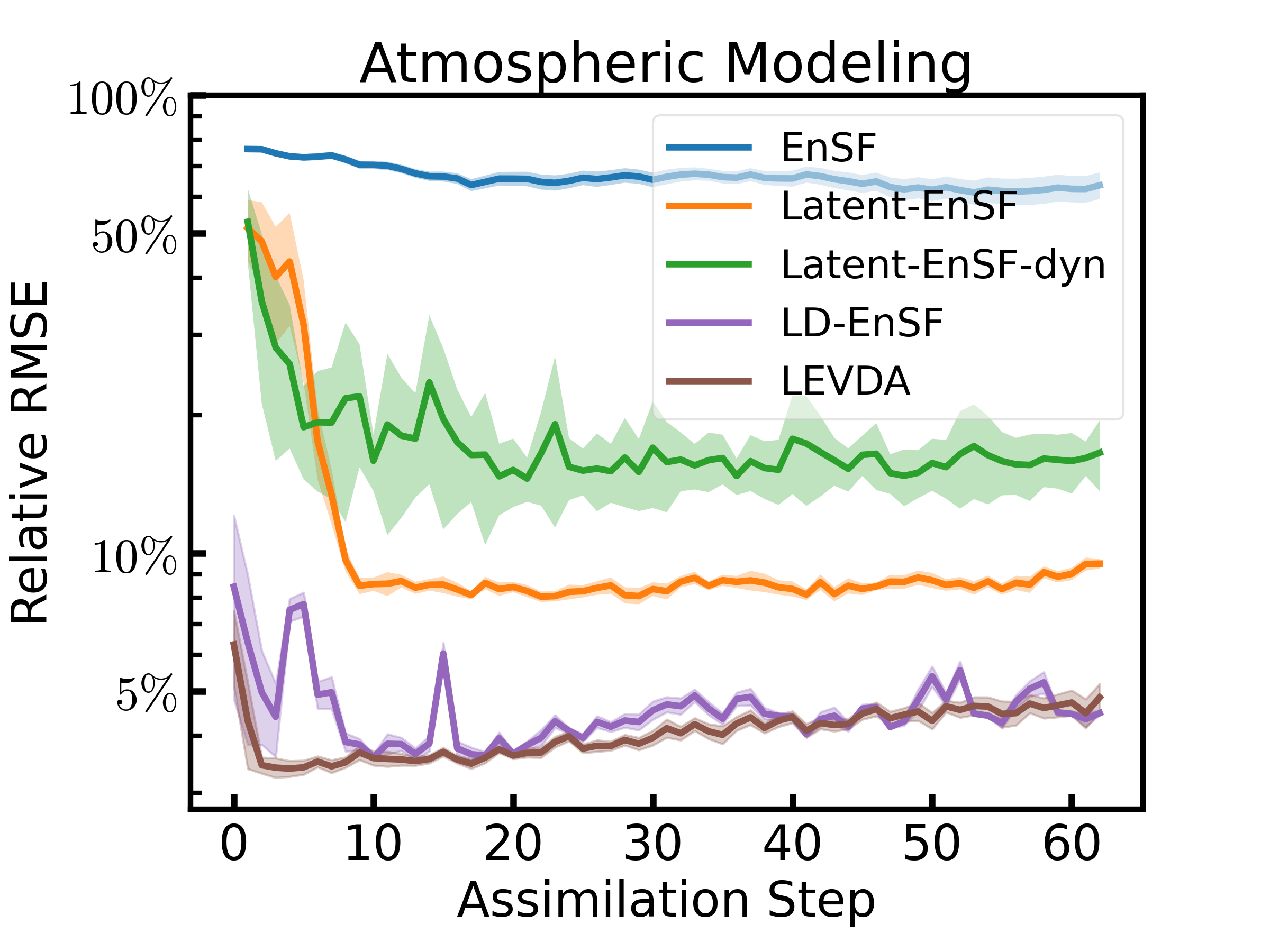}
    \caption{Relative RMSE of the ensemble-mean trajectory over time for Kolmogorov flow (left), tsunami modeling (middle), and atmospheric modeling (right); shaded regions show the standard deviation of member-wise relative RMSE.}
	  \label{fig:main_results}
	  \vspace{-10pt}
\end{figure*}

Figure~\ref{fig:main_results} summarizes assimilation accuracy and uncertainty across the three benchmarks. Overall, LEVDA matches or outperforms LD-EnSF without requiring an auxiliary LSTM observation encoder, and it consistently surpasses the remaining baselines under sparse observations. Additionally, Appendix~\ref{app:ablation} reports that LEVDA achieves more robust assimilation with varying assimilation window length $\tau$ and the ensemble size $K$, varying levels of observation sparsity and noise, as well as more accurate parameter estimation. Improved uncertainty quantification measured by the spread-error ratio and energy score for LEVDA is reported in Appendix~\ref{app:uq}.

\begin{figure*}[t]
  \centering
    \includegraphics[width=0.95\textwidth]{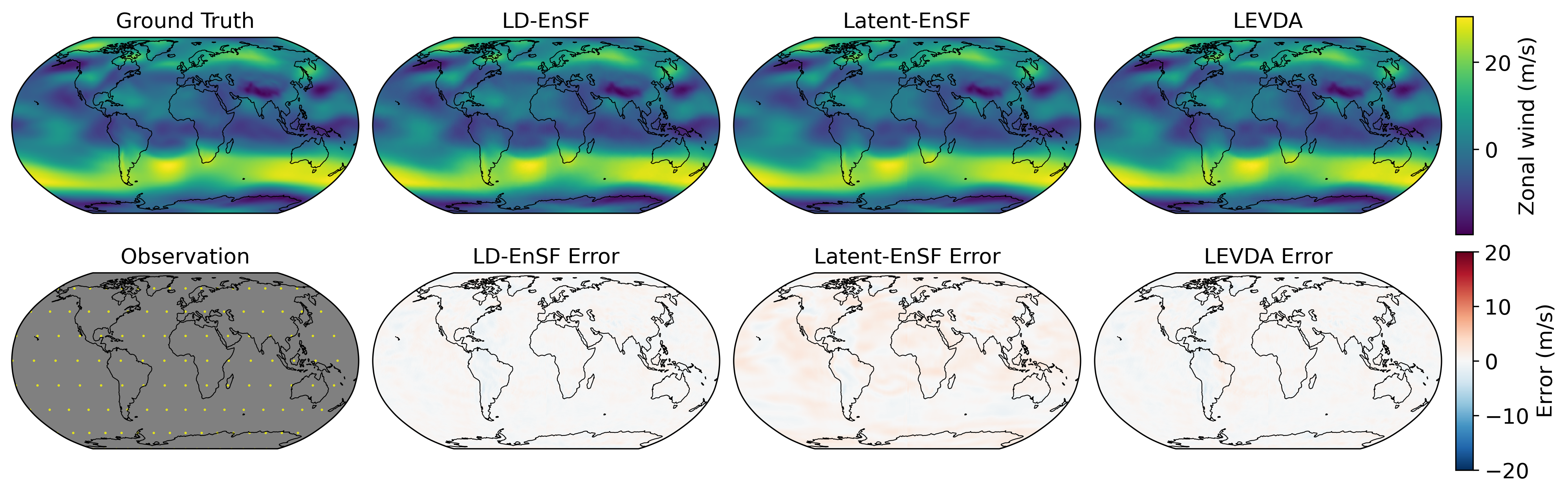}
  \caption{Atmospheric benchmark: final-time zonal wind $u$ at 500\,hPa. The visualization compares the ground truth (with sparse observations) to the assimilated states and errors for different methods.}
	  \label{fig:planetswe_visual}
\end{figure*}

Figure~\ref{fig:planetswe_visual} visualizes the final-time zonal wind velocity at 500\,hPa for the atmospheric benchmark. Despite the sparse observation density, gradient propagation through the latent surrogate remains stable, allowing LEVDA to produce realistic reconstructions comparable to LD-EnSF. Extended temporal visualizations comparing assimilation performance for Kolmogorov flow and tsunami modeling are provided in Appendix~\ref{sec:appendix_viz}.

\subsection{Spatiotemporally Varying Observation Operators}
\begin{figure*}[h]
  \centering
  \begin{subfigure}[t]{0.33\linewidth}
     \centering
    \includegraphics[width=\linewidth]{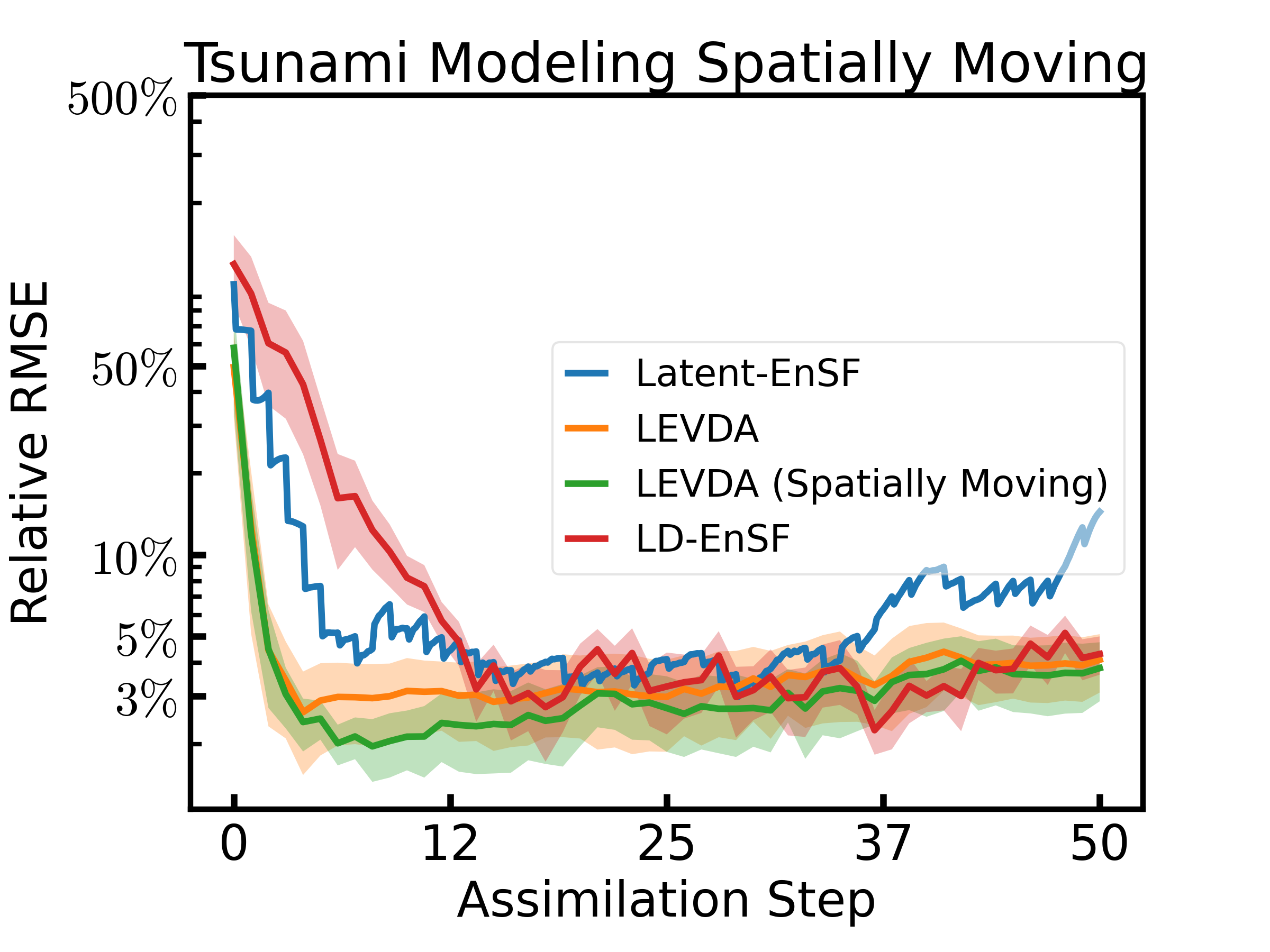}
  \end{subfigure}\hspace{-0.2em}
  \begin{subfigure}[t]{0.33\linewidth}
     \centering
    \includegraphics[width=\linewidth]{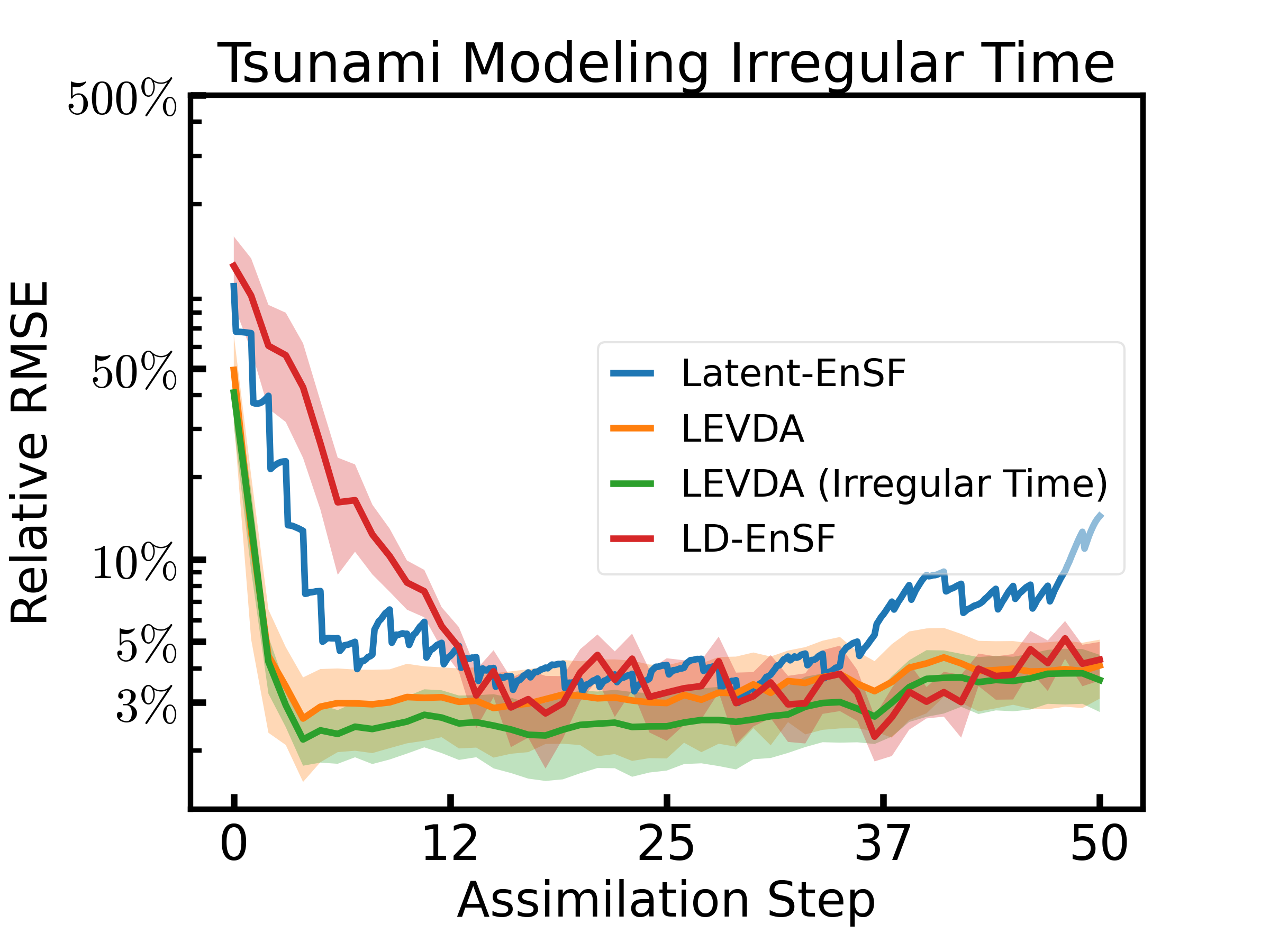}
  \end{subfigure}\hspace{-0.2em}
  \begin{subfigure}[t]{0.33\linewidth}
     \centering
    \includegraphics[width=\linewidth]{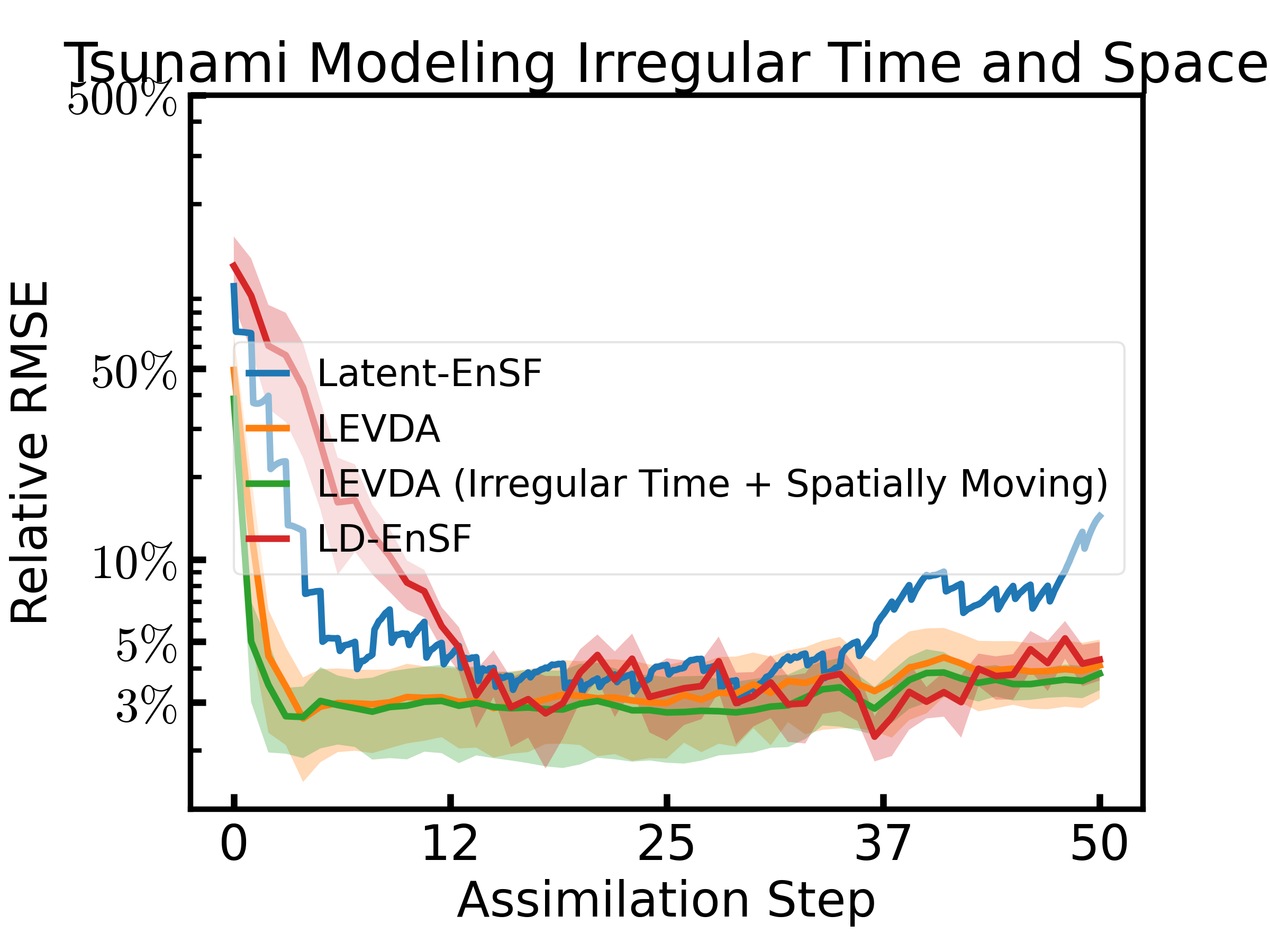}
  \end{subfigure}
  \caption{Effect of irregular sampling on tsunami modeling: (a) moving observation locations (randomized at each observed time), (b) irregular observation times with fixed spatial locations, and (c) joint spatiotemporal irregularity. Curves depict the relative RMSE of the ensemble-mean trajectory; shaded regions indicate the standard deviation of member-wise relative RMSE.}
  \label{fig:ablation_irregular_obs}
  \vspace{-10pt}
\end{figure*}

LEVDA is designed to handle flexible observation patterns, relaxing the standard assumption that measurements must arrive on fixed grids at regular intervals. In practice, data is often collected by moving platforms or sampled asynchronously. We evaluate LEVDA under three progressively more realistic scenarios: time-varying observation locations, irregular observation timestamps, and a joint regime combining both irregularities. As shown in Figure~\ref{fig:ablation_irregular_obs}, LEVDA maintains performance comparable to the idealized fixed-grid baseline in all cases.

\textbf{Moving Observation Locations}.
LEVDA overcomes the fixed-grid limitation of encoder-based methods like LD-EnSF, enabling assimilation from time-varying locations typical of moving platforms or drifting sensors. We evaluate this using the tsunami benchmark (Table~\ref{tab:exp-settings}), comparing the standard fixed-grid setup to a scenario where observation coordinates are resampled at each time step (maintaining $0.4\%$ sparsity). Figure~\ref{fig:ablation_irregular_obs}a shows that LEVDA adapts robustly to these spatially shifting observations, achieving accuracy comparable to the fixed-grid baseline.

\textbf{Irregular Observation Times}.
LDNets model the system dynamics via a learned time derivative of the latent state, rather than a fixed-step recurrence. This continuous-time formulation allows the latent state to be propagated using standard numerical integrators (e.g., forward Euler) to any arbitrary timestamp. Consequently, LEVDA can naturally assimilate observations taken at irregular or asynchronous intervals, decoupling the assimilation process from the rigid fixed-time grids required by many discrete-time data-driven models.

In Figure~\ref{fig:ablation_irregular_obs}b, we examine an irregular-sampling regime by drawing 51 observation times uniformly at random, matching the total count of the fixed-interval baseline. Observation times that do not fall at regular time steps are calculated by (\ref{eq:interp}).

During smoothing, LEVDA assimilates only those observations that fall within the current time window; consequently, the number of available observations per window varies naturally. For valid comparison, all methods are evaluated on the original fixed simulation grid, even though LEVDA (Irregular Time) assimilates the asynchronously sampled data.

\textbf{Joint Spatiotemporal Irregularity}.
We combine spatial subsampling and temporal irregularity to form a joint spatiotemporal stress test. At each randomly selected observation time, we independently draw a random subset of spatial locations. Consistent with the findings in Figure~\ref{fig:ablation_irregular_obs}b, Figure~\ref{fig:ablation_irregular_obs}c demonstrates that LEVDA maintains performance comparable to the fixed-interval baseline even under this fully irregular regime.

\subsection{Runtime and Training Cost}
Table~\ref{tab:computation_time} compares wall-clock runtimes for 4DEnVar, Latent-EnSF, LD-EnSF, and LEVDA. LEVDA eliminates the need to train an observation-to-latent encoder (as required by LD-EnSF) and avoids the prohibitive cost of full-state 4DEnVar. Instead, it relies on iterative ensemble-space optimization, utilizing gradients computed via automatic differentiation through the latent surrogate. Overall, LEVDA remains significantly faster than the full-dimensional 4DEnVar baseline while maintaining competitive accuracy under sparse observations.

\begin{table*}[h] 
	    \setlength{\tabcolsep}{4pt}
	   \resizebox{\textwidth}{!}{\begin{tabular}{c|cccc|cccc|cccc}
    \toprule
    \centering
        Example & \multicolumn{4}{c|}{Kolmogorov Flow} & \multicolumn{4}{c|}{Tsunami Modeling} & \multicolumn{4}{c}{Atmospheric Modeling} \\ \midrule
        \textbf{Metric} & 4DEnVar & Latent-EnSF & LD-EnSF &  LEVDA
& 4DEnVar & Latent-EnSF & LD-EnSF &  LEVDA & 4DEnVar & Latent-EnSF & LD-EnSF &  LEVDA \\ \midrule
        Encoder Training (s) & $0$ & $43,156$ & $298$ & $0$ & $0$ & $27,718$ & $312$ & $0$ & $0$ & $58,231$ & $1429$ & $0$\\
        Assimilation (s) & $-$ & $7.57$ & $0.35$ & $13.91$ & $24,639$ & $ 5.77$ & $0.37$ & $11.32$ & $-$ & $ 37.5$ & $1.42$ & $19.47$\\
        Assimilation dim. & $D_x{=}45{,}000$ &$D_s{=}499$ & $D_s{=}10$ & $D_s{=}10$& $D_x{=}67{,}500$ & $D_s{=}400$ & $D_s{=}12$ & $D_s{=}12$ & $D_x{=}393{,}216$ & $D_s{=}512$ & $D_s{=}52$ & $D_s{=}52$\\ 
        \bottomrule
    \end{tabular}}
    \caption{Wall-clock runtime comparison. We report the latent assimilation dimension $D_s$ for latent methods (Latent-EnSF, LD-EnSF, LEVDA) and the full-state dimension $D_x$ for 4DEnVar. Assimilation runtimes correspond to full-trajectory assimilation with an ensemble size of $K=100$ for latent methods, whereas 4DEnVar is restricted to $K=40$ due to the high computational cost of full-state propagation. 4DEnVar simulations use an AMD 7543 CPU (64 threads), while data-driven and latent methods utilize an Nvidia A6000 GPU.}
    \label{tab:computation_time}
\vspace{-10pt}
\end{table*}
	
\section{Related Work}

We position our work at the intersection of (i) latent-space filtering for sparse observations, (ii) score-based smoothing via conditional generation, and (iii) surrogate-accelerated variational data assimilation. Latent-EnSF \citep{si2025latent} and LD-EnSF \citep{xiao2026ld} demonstrate that score-based priors in a low-dimensional latent space render \emph{filtering} feasible for large-scale systems with sparse observations. However, as sequential methods, they enforce weaker trajectory-level constraints across multi-time windows. Additionally, they often require training auxiliary components (e.g., observation-to-latent encoders) to map measurements into the latent space.

In contrast, score-based \emph{smoothing} methods such as SDA \citep{rozet2023score} and FlowDAS \citep{chen2025flowdas} target the smoothing posterior via conditional trajectory generation. While these approaches avoid explicit adjoint derivations, their inference cost is often dominated by the large number of required denoising or flow steps. Moreover, they typically represent dynamics implicitly through a generative model, rather than via an explicit differentiable surrogate that supports efficient gradient-based windowed optimization and joint state--parameter inference. Similarly, although APPA \citep{andry2025appa} learns dynamics via generative models, extending it to handle joint parameter inference and irregular sampling would require non-trivial architectural changes.

Finally, data-driven surrogates are increasingly replacing expensive PDE solvers to accelerate assimilation. For instance, FengWu-4DVar \citep{xiao2024fengwu} employs a fast surrogate for single-trajectory 4DVar but yields a MAP estimate \citep{hodyss2016extent}, providing limited distributional guidance. LEVDA instead optimizes within a latent ensemble subspace, naturally preserving ensemble outputs for uncertainty quantification. A detailed qualitative comparison is provided in Appendix \ref{app:capability}.

\section{Conclusion}
We introduced Latent Ensemble Variational Data Assimilation (LEVDA), an ensemble-variational smoother that performs 4DEnVar-style optimization in the low-dimensional latent space of a pretrained differentiable dynamical surrogate. By optimizing over an ensemble subspace and differentiating through the surrogate via automatic differentiation, LEVDA enables efficient windowed assimilation while bypassing the need for tangent-linear/adjoint implementations of the high-dimensional simulator or auxiliary observation-to-latent encoders. LEVDA jointly assimilates states and unknown parameters, naturally accommodates time-dependent observation operators (including moving sensors), and supports observations at arbitrary spatiotemporal locations. Across three challenging geophysical benchmarks, LEVDA matches or outperforms state-of-the-art latent filtering baselines under severe sparsity while providing substantial accuracy and runtime advantages over full-state 4DEnVar. Finally, our ablations demonstrate robustness to spatiotemporal irregularity and modest ensemble sizes.

\textbf{Limitations.} LEVDA relies on a pretrained differentiable surrogate and inherits its potential failure modes: assimilation accuracy depends on the surrogate's fidelity over the window, and performance may degrade under significant distribution shifts or unmodeled model errors. Furthermore, LEVDA optimizes a nonconvex objective within an ensemble subspace. While the low latent dimension ensures tractability and Appendix~\ref{app:ablation} demonstrates robustness across varying $\tau$, $K$, and sparsity levels, difficult regimes may still require careful tuning of hyperparameters such as ensemble size and inflation. Finally, although LEVDA improves uncertainty quantification, achieving consistent calibration, especially for atmospheric dynamics, remains a challenge. Future work incorporating explicit weak-constraint model-error terms or adaptive regularization offers a promising path forward.

\section{Acknowledgments}

This work was supported in part by the National Science Foundation under Grant No. 2325631. We thank Pengpeng Xiao for her collaboration on our work LD-EnSF~\citep{xiao2026ld}, which provided part of the LDNet implementation and benchmarks for this project.

\bibliographystyle{icml2026}
\bibliography{references}

\newpage    
\appendix

\section{Ablation Studies on Tsunami Modeling}\label{app:ablation}
\begin{figure*}[!h]
  \centering
  \begin{subfigure}[t]{0.33\linewidth}
     \centering
    \includegraphics[width=\linewidth]{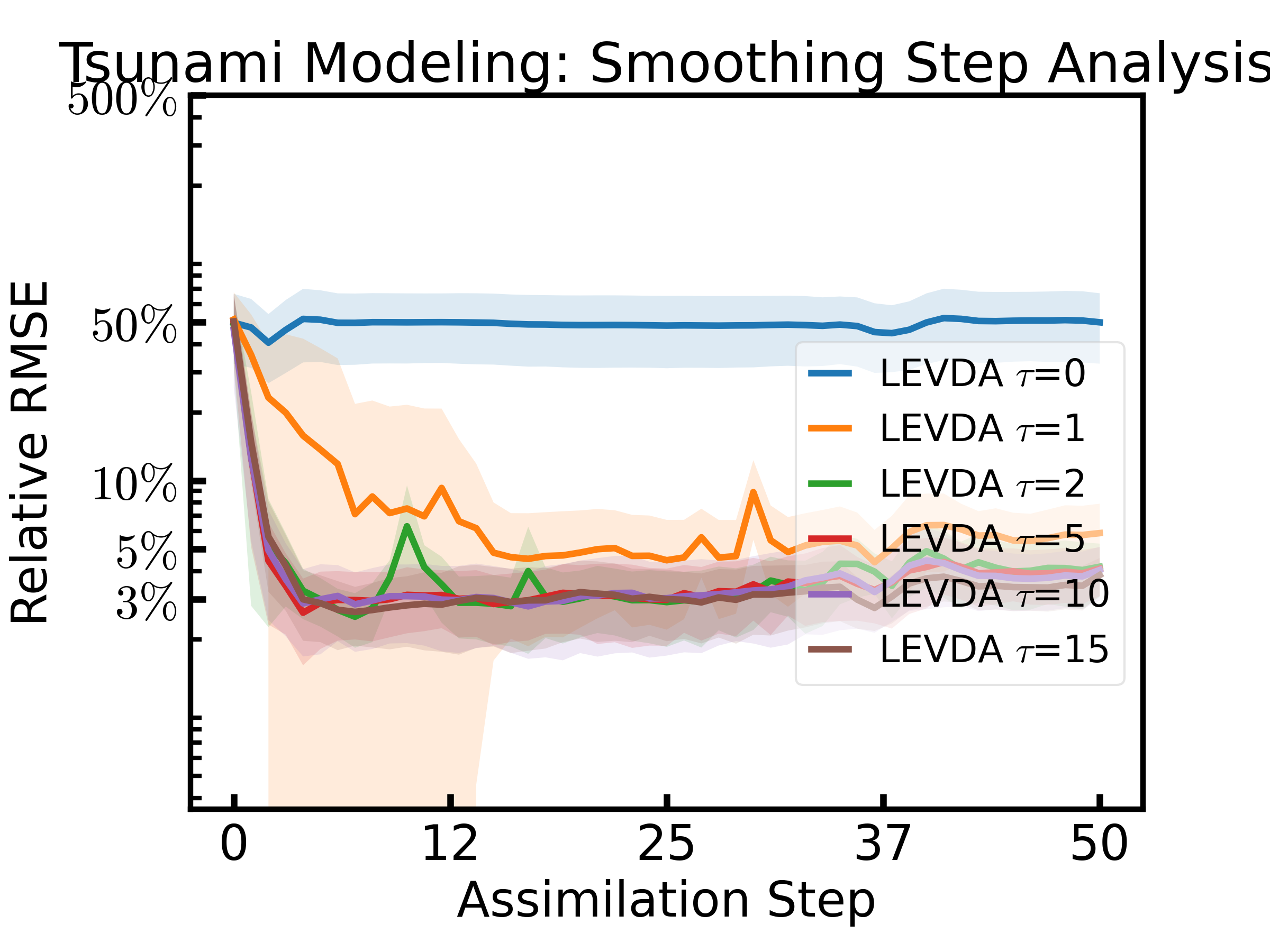}
  \end{subfigure}\hspace{-0.2em}
  \begin{subfigure}[t]{0.33\linewidth}
     \centering
    \includegraphics[width=\linewidth]{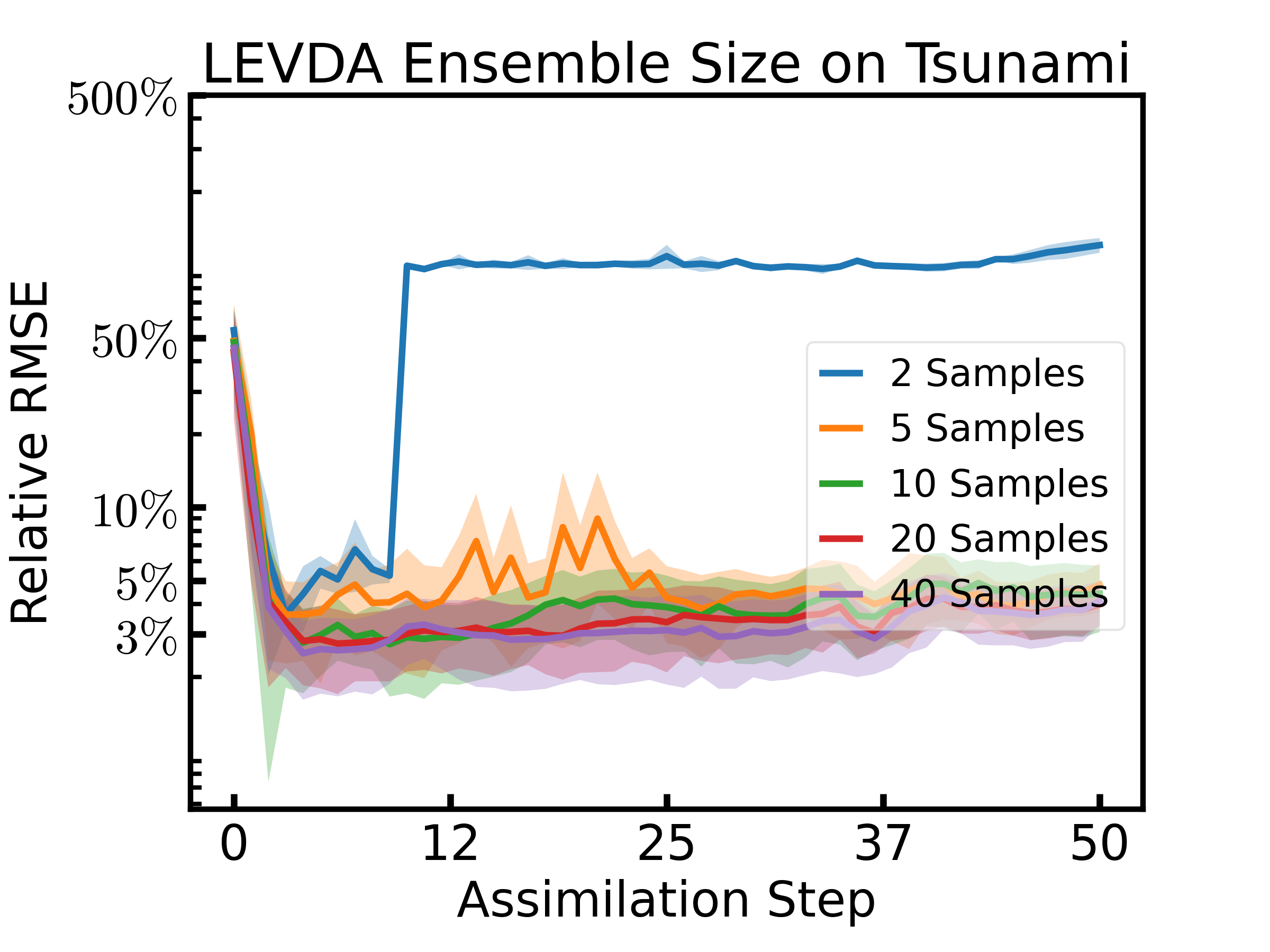}
  \end{subfigure}\hspace{-0.2em}
  \begin{subfigure}[t]{0.33\linewidth}
     \centering
    \includegraphics[width=\linewidth]{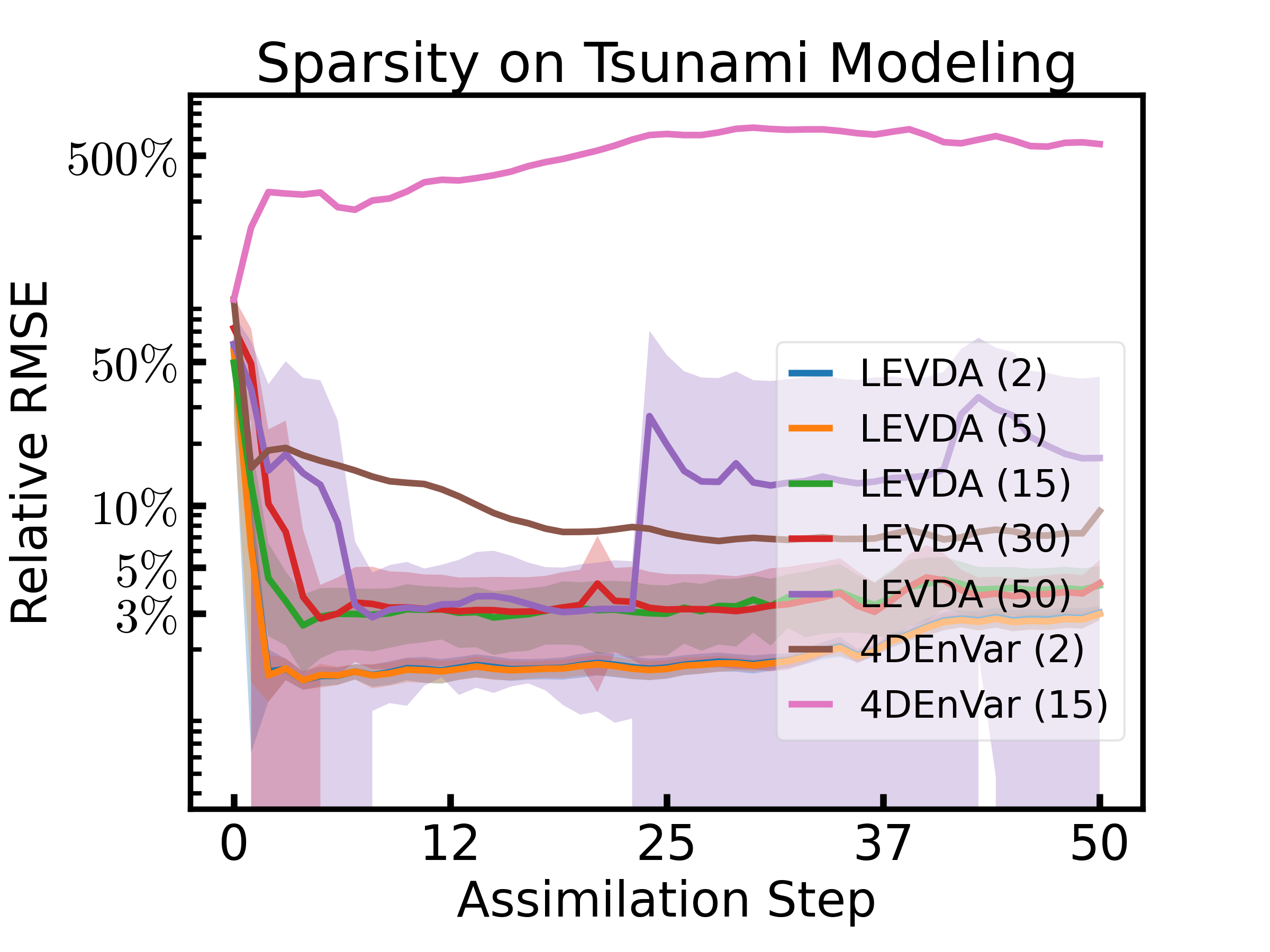}
  \end{subfigure}

\caption{Tsunami benchmark sensitivity analysis: LEVDA Relative RMSE as a function of (a) smoothing time horizon $\tau$, (b) ensemble size $K$, and (c) spatial observation stride (sparsity).}
\label{fig:ablation_studies}
\end{figure*}

We perform ablation studies using the tsunami modeling benchmark to evaluate LEVDA's robustness to three key hyperparameters: smoothing time horizon, ensemble size, and observation sparsity. Figure~\ref{fig:ablation_studies} illustrates the sensitivity of the reconstruction error to variations in these settings.

\textbf{Smoothing Time Horizon.}
We vary the smoothing time horizon $\tau$ (the length of the assimilation window) to assess robustness to the duration of temporal context utilized.
Figure~\ref{fig:ablation_studies}a demonstrates that LEVDA is relatively insensitive to $\tau$ beyond a critical threshold.
The failure observed at $\tau=0$ is consistent with theoretical expectations: an instantaneous snapshot provides insufficient dynamic information to uniquely identify the governing parameters $u$.

\textbf{Ensemble Size.}
We vary the ensemble size $K$ to evaluate the sample efficiency of the data assimilation process.
Figure~\ref{fig:ablation_studies}b indicates that performance stabilizes for $K \ge 10$, with moderate degradation at $K=5$ and divergence at $K=2$.
This efficiency stems from performing optimization within a low-dimensional latent subspace, where a relatively small ensemble is sufficient to span the tangent space required for effective updates.

\textbf{Spatial Observation Sparsity.}
To evaluate robustness to data scarcity, we test spatial strides $s \in \{2,5,15,30,50\}$, corresponding to observation densities ranging from $25\%$ down to an extreme $0.04\%$ on the 2D grid.
We benchmark against 4DEnVar (with localization radius $r=2$) at strides $2$ and $15$.
Figure~\ref{fig:ablation_studies}c reveals a stark contrast: while 4DEnVar is competitive at high density ($s=2$), it suffers a performance collapse at the experimental setting of $s=15$ ($0.4\%$).
Conversely, LEVDA maintains high accuracy and stability up to $s=15$, degrades only modestly at $s=30$, and remains functional even at the hyper-sparse rate of $0.04\%$ ($s=50$).

Our experiments reveal that 4DEnVar is highly sensitive to the localization radius under sparse conditions.
Extreme sparsity creates a fundamental dilemma: a large localization radius introduces sampling noise via spurious long-range correlations, causing filter divergence, whereas a small radius prevents the propagation of information from sparse observations to the vast unobserved regions.
At $s=15$ ($0.4\%$ density), we observe a specific failure mode where the analysis fits the observations well (low residual) but fails globally (high state error). 
This occurs because the localization mask truncates the physical cross-covariances necessary to infer the unobserved state variables, rendering the full-state inverse problem ill-posed, a limitation LEVDA avoids by assimilating in a globally coupled latent space.
 
\begin{figure*}[!h]
  \centering
  \begin{subfigure}[t]{0.33\linewidth}
     \centering
    \includegraphics[width=\linewidth]{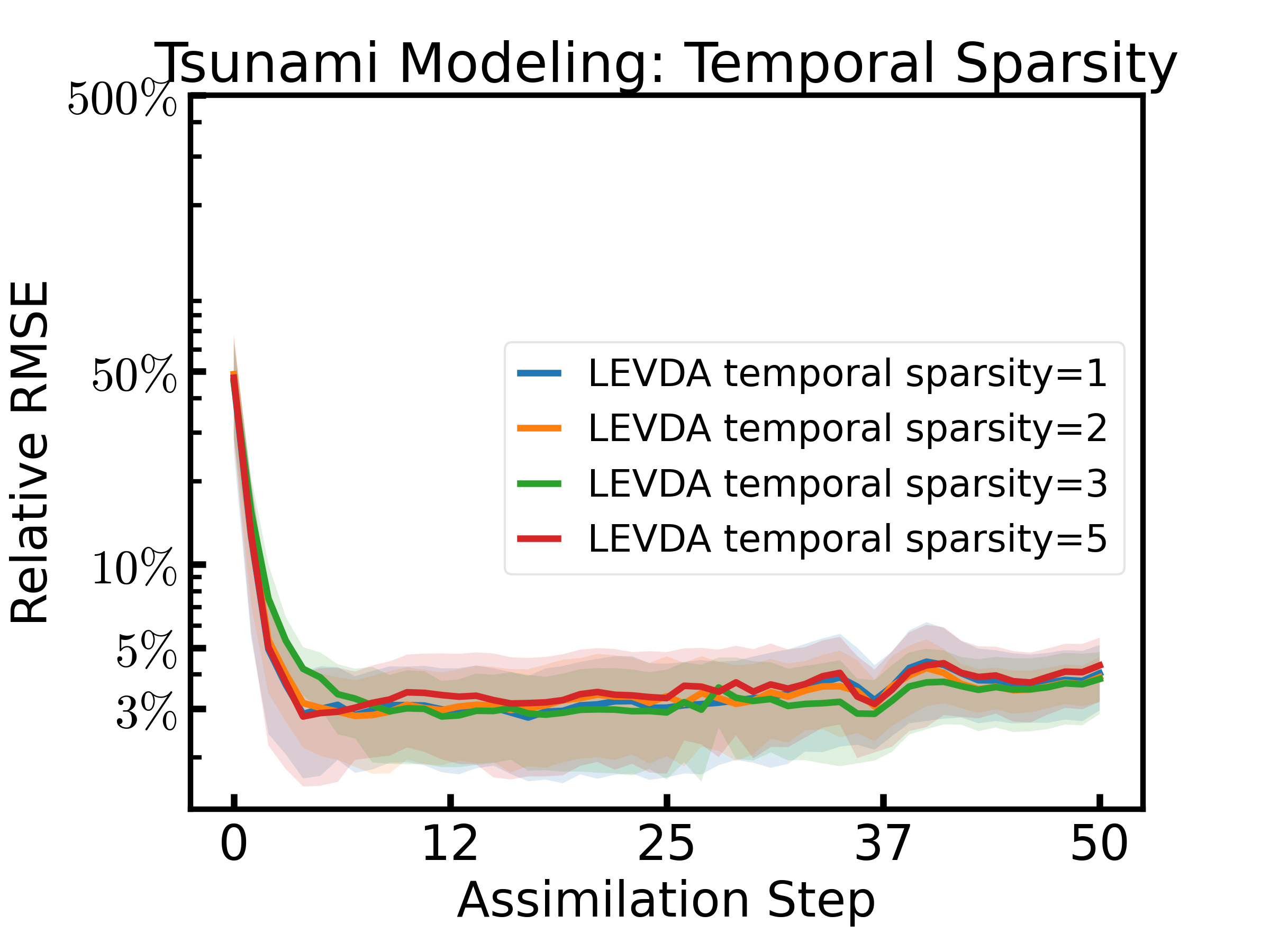}
  \end{subfigure}\hspace{-0.2em}
  \begin{subfigure}[t]{0.33\linewidth}
     \centering
    \includegraphics[width=\linewidth]{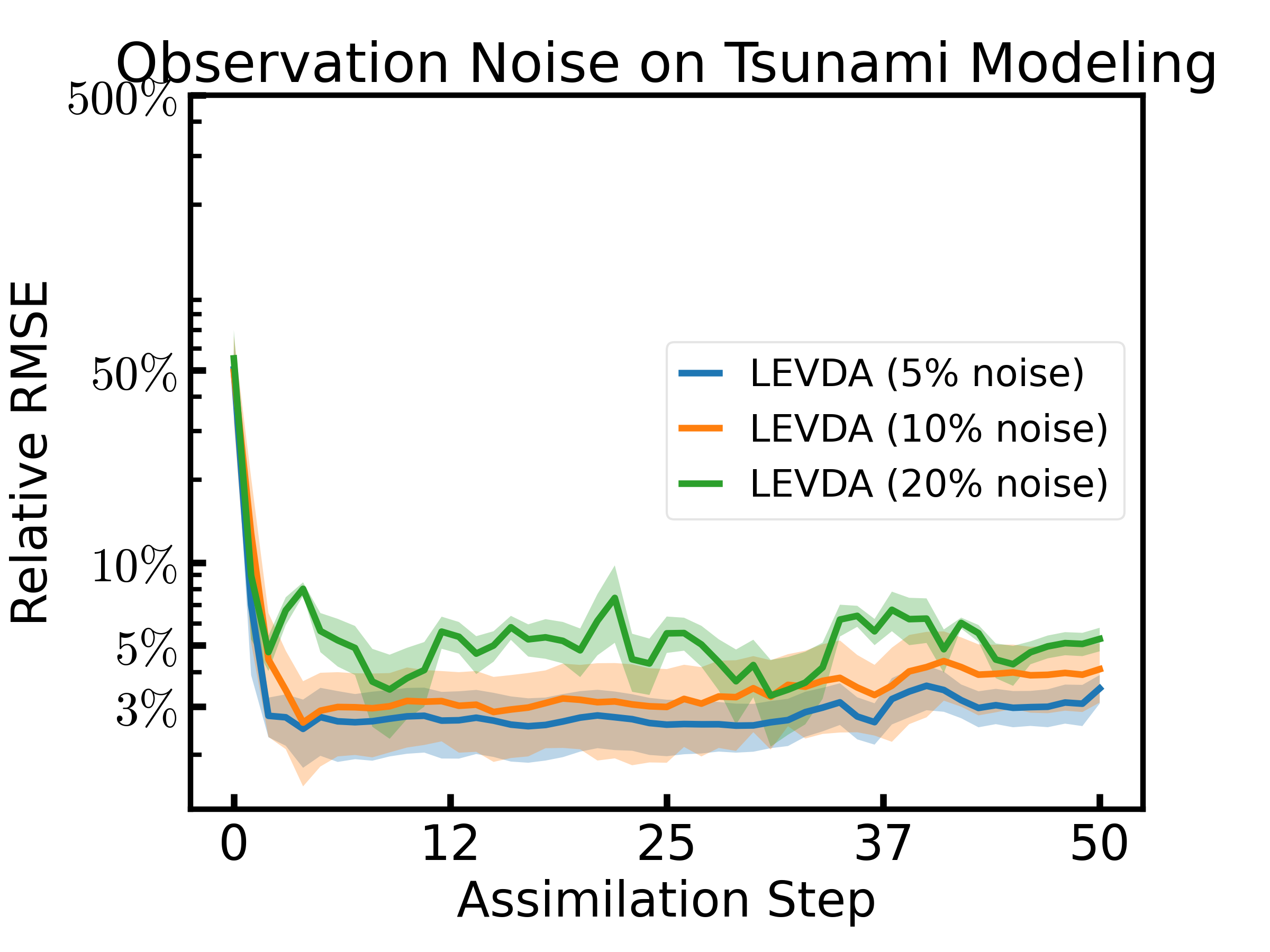}
  \end{subfigure}\hspace{-0.2em}
  \begin{subfigure}[t]{0.33\linewidth}
     \centering
    \includegraphics[width=\linewidth]{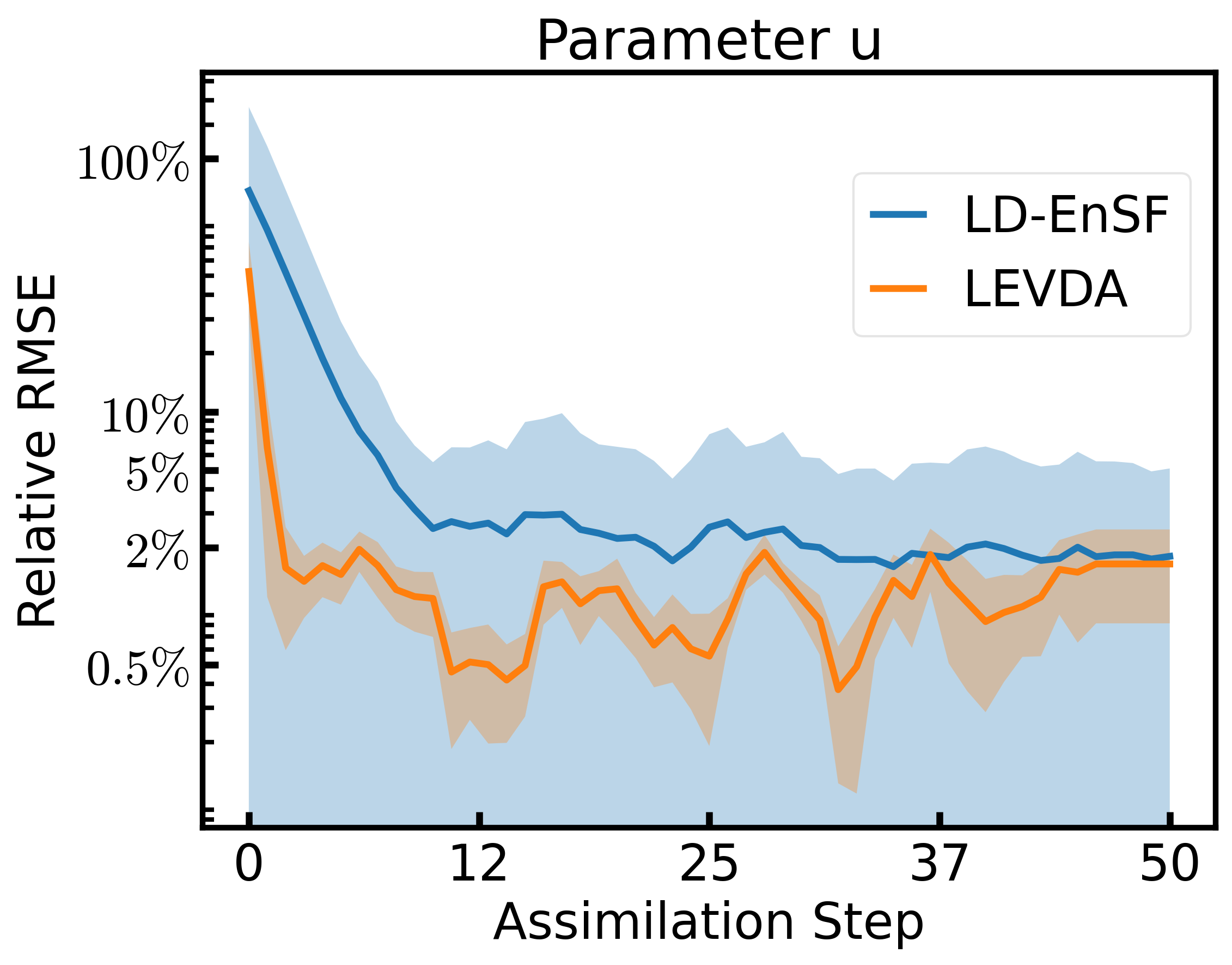}
  \end{subfigure}

  \caption{Additional Tsunami benchmark ablations: (a) Impact of temporal sparsity on LEVDA Relative RMSE (using extended smoothing window $\tau=10$). (b) Robustness to observation noise (5\%, 10\%, 20\% noise-to-signal ratios) under fixed likelihood weighting. (c) Parameter estimation accuracy: Relative RMSE of the parameter $u$ over the assimilation window.}
  \label{fig:ablation_studies_2}
\end{figure*}

\textbf{Robustness to Temporal Sparsity.}
We examine LEVDA's sensitivity to temporal data scarcity by increasing the time interval between observations.
Figure~\ref{fig:ablation_studies_2}a demonstrates remarkable stability: error rates remain nearly flat even as observations become increasingly infrequent.
This resilience stems from optimizing a continuous trajectory within the low-dimensional latent manifold, which effectively bridges the gaps between sparse timepoints.

For this experiment, we extended the smoothing window to $\tau=10$ (versus the standard $\tau=5$) to ensure the assimilation horizon remains sufficiently long to capture these wider observation gaps. 

\textbf{Sensitivity to Observation Noise.}
We analyze LEVDA's robustness to varying magnitudes of observation noise, addressing the reality that real-world sensor statistics are often uncertain.
Because the noise level dictates the balance between the observation likelihood and the learned latent prior, the assimilation framework must remain stable even when this balance shifts.
Figure \ref{fig:ablation_studies_2}b reports performance under noise-to-signal ratios of 5\%, 10\%, and 20\% for the tsunami modeling case.
Crucially, we maintain identical hyperparameter weights across all trials, demonstrating that LEVDA handles increased measurement uncertainty without requiring specific retuning.

\textbf{Parameter Estimation.}
Beyond state reconstruction, many inverse problems necessitate the recovery of governing latent parameters.
In the tsunami modeling benchmark, we focus on estimating the initial condition parameter $u$ that specifies the starting location of the Gaussian bump.
Although $u$ is not directly observed, it dictates the subsequent wave evolution and must be inferred solely from sparse partial observations.
Figure \ref{fig:ablation_studies_2}c tracks the RMSE of the estimated parameter over the course of the assimilation window.
LEVDA consistently achieves a lower RMSE than LD-EnSF, despite not utilizing an explicit encoder as in LD-EnSF to map observations to parameters.
Instead, LEVDA treats $u$ as an augmented component of the latent state, estimating it jointly via the variational smoothing objective.
This approach leverages the full temporal context of the smoothing window, allowing observations from multiple time steps to progressively constrain $u$, thereby reducing the uncertainty often associated with purely filtering-based updates.
The superior accuracy of LEVDA confirms that variational smoothing offers a more robust mechanism for latent parameter discovery than direct encoder-based inversion.

\section{Uncertainty Quantification Metrics}\label{app:uq}

We evaluate the quality of LEVDA’s uncertainty estimates using proper scoring rules and calibration diagnostics. Across all test cases, LEVDA consistently improves these metrics, demonstrating a more reliable characterization of ensemble dispersion. However, achieving proper calibration in the atmospheric modeling case remains a significant challenge for all evaluated methods.

\begin{figure*}[!htb]
    \centering
    \includegraphics[width=0.33\linewidth]{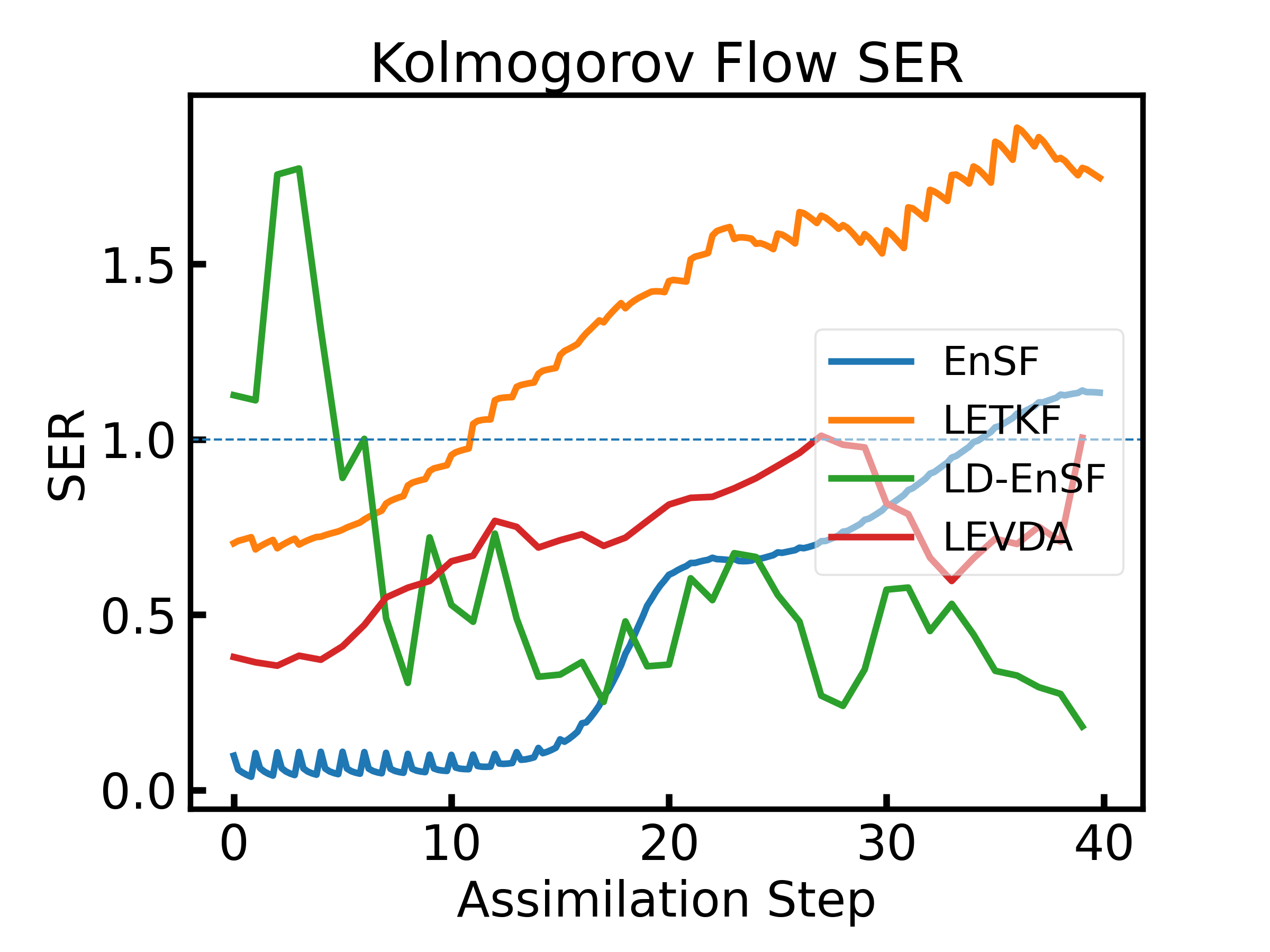}\hspace{-0.2em}
    \includegraphics[width=0.33\linewidth]{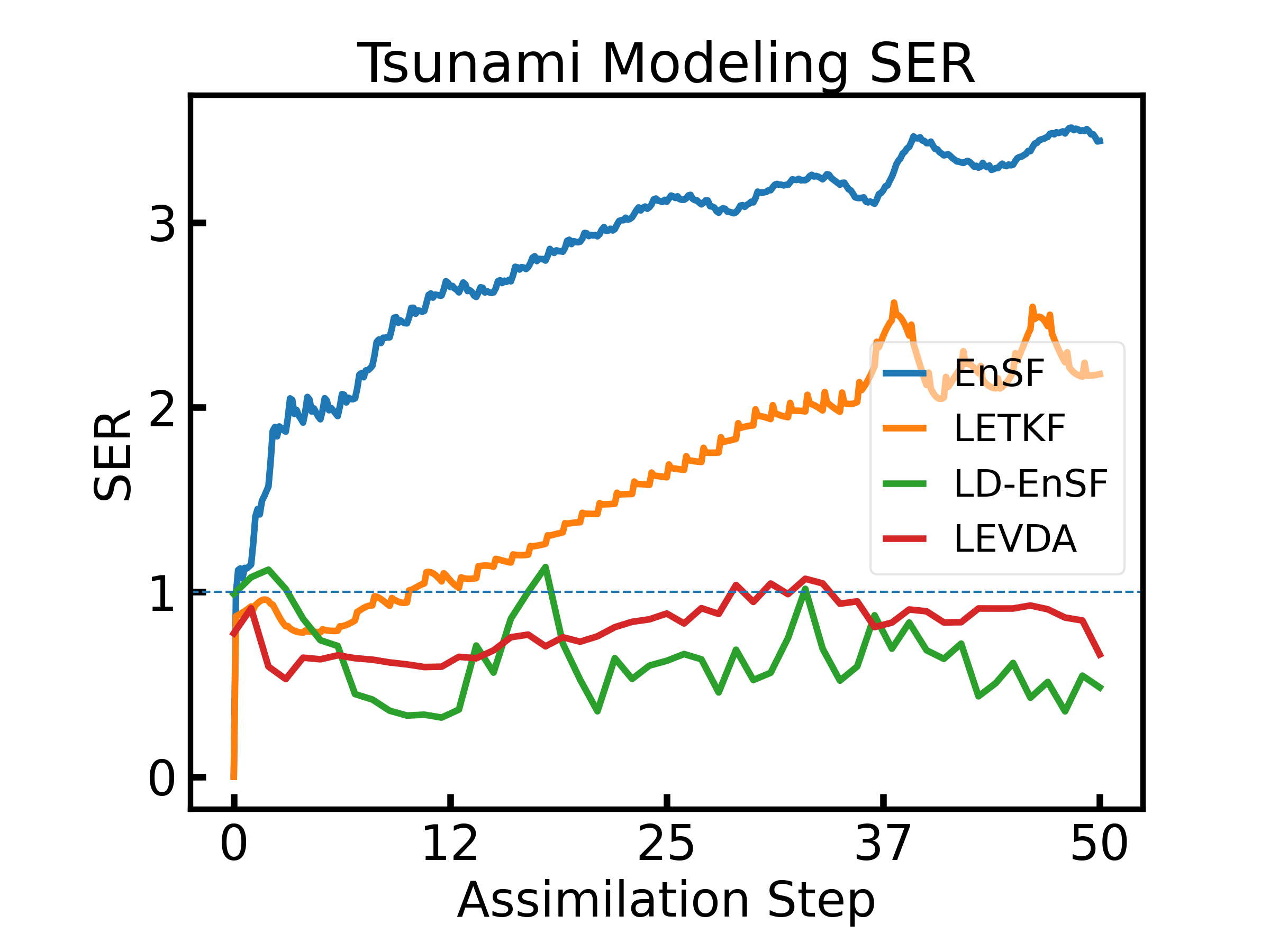}\hspace{-0.2em}
    \includegraphics[width=0.33\linewidth]{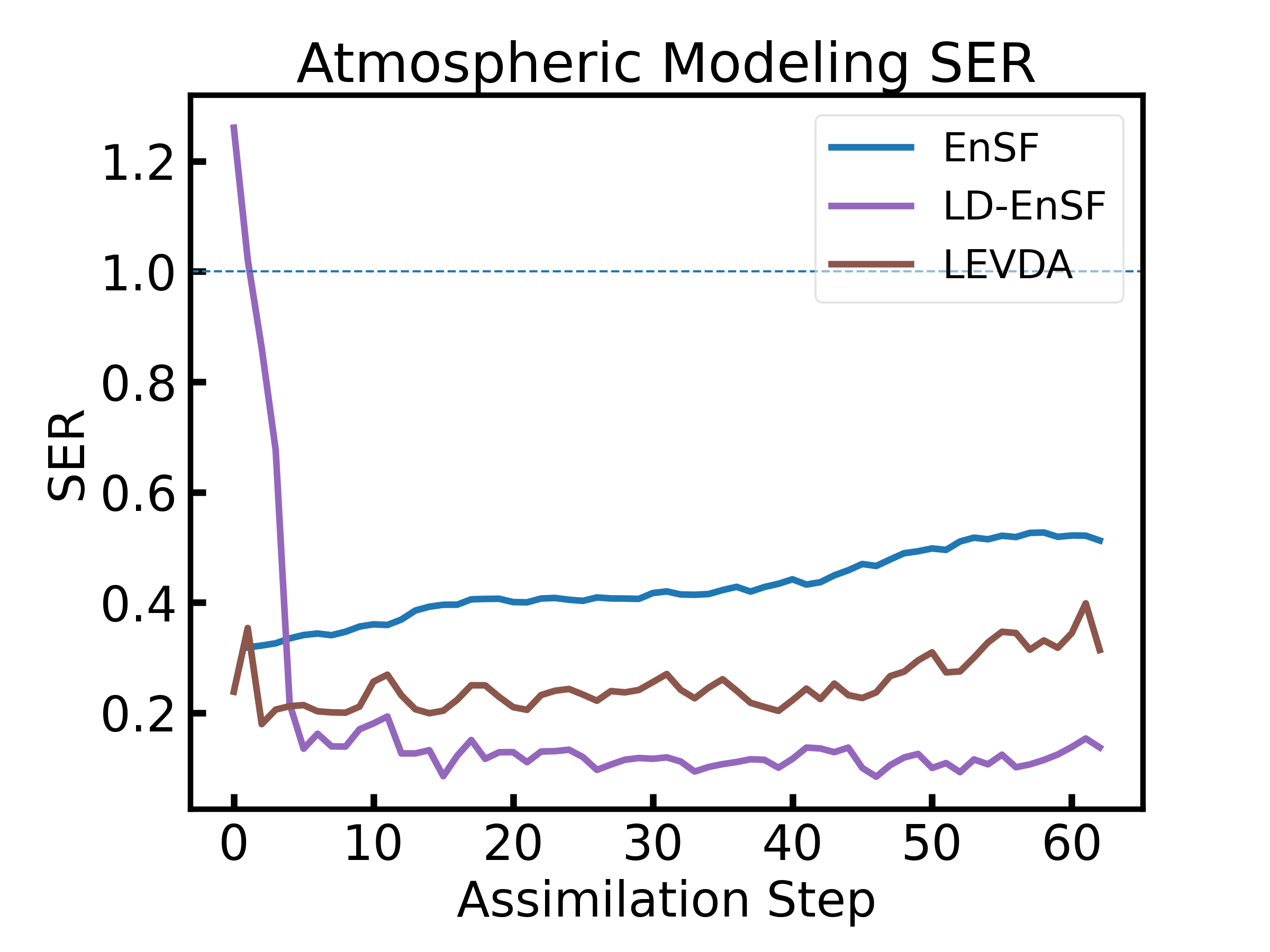}
    \caption{Spread-Error Ratio (SER) over time for Kolmogorov flow, tsunami modeling, and atmospheric modeling (left-to-right). Values closer to $1$ indicate better calibration of the ensemble spread.}
	  \label{fig:ser_main}
	  \vspace{-10pt}
\end{figure*}

\textbf{Spread-Error Ratio.}
To evaluate the reliability of the assimilated ensemble, we first employ the Spread-Error Ratio (SER). This metric diagnoses the calibration of the ensemble by comparing its internal diversity (spread) to the actual predictive accuracy of its mean.
The SER at time $t$ is defined as:
\begin{align}
\text{SER}_t =
\frac{\mathrm{Spread}_t}{\mathrm{Error}_t}
=
\frac{
\sqrt{\frac{1}{K}\sum_{j=1}^K \lVert x_t^{(j)} - \bar{x}_t\rVert_2^2}
}{
\lVert \bar{x}_t - x_t^\star\rVert_2
},
\end{align}
where $\bar{x}_t = \frac{1}{K}\sum_{j=1}^K x_t^{(j)}$ is the ensemble mean, $x_t^{(j)}$ represents individual ensemble members, and $x_t^\star$ is the ground truth. 
An SER of 1 indicates ideal calibration, whereas SER $<$ 1 signals overconfident under-dispersion, and SER $>$ 1 signals underconfident over-dispersion. We report the evolution of the SER across the three benchmarks in Figure \ref{fig:ser_main}.

\begin{figure*}[!htb]
\centering
\includegraphics[width=0.33\linewidth]{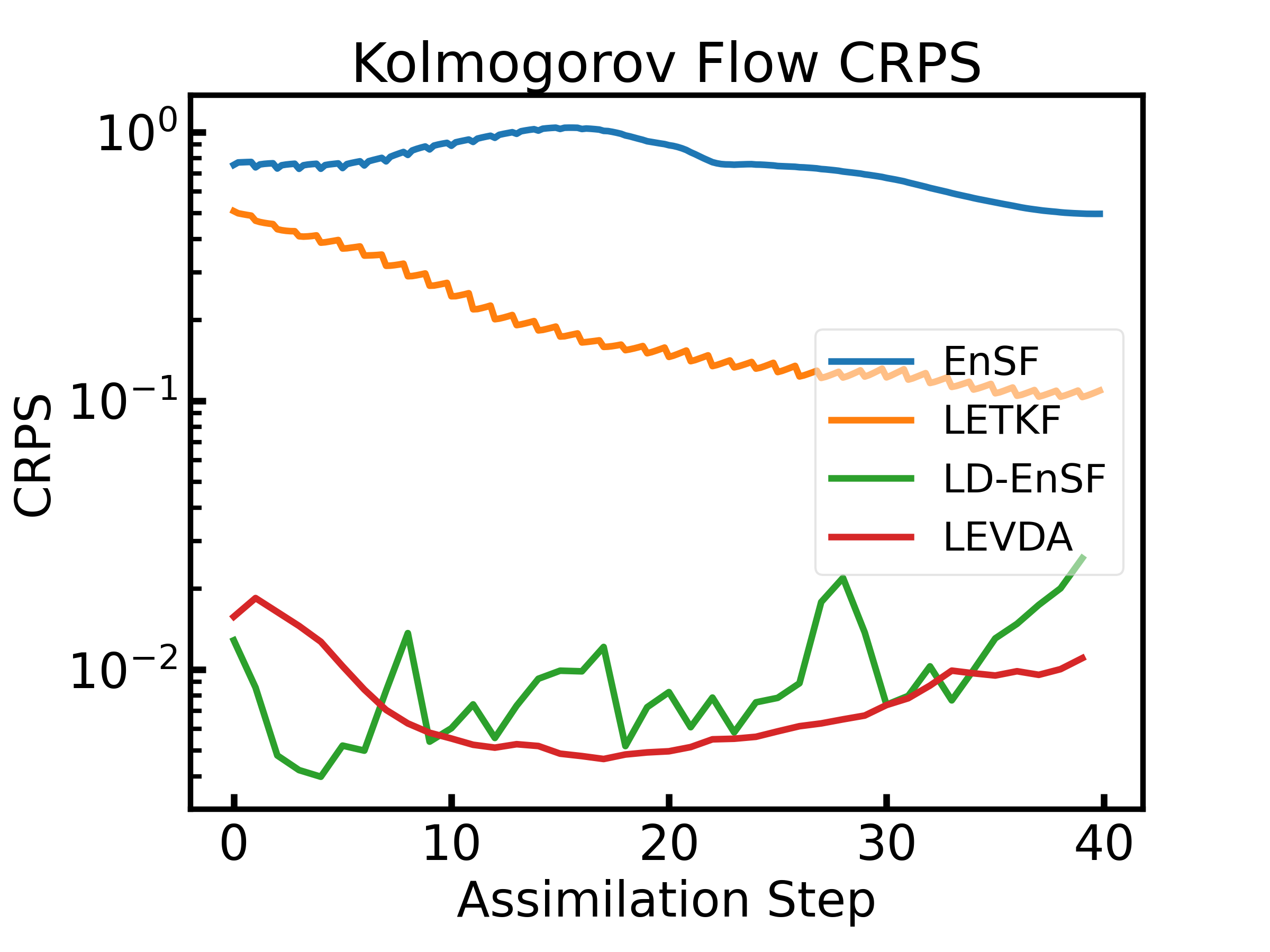}\hspace{-0.2em}
\includegraphics[width=0.33\linewidth]{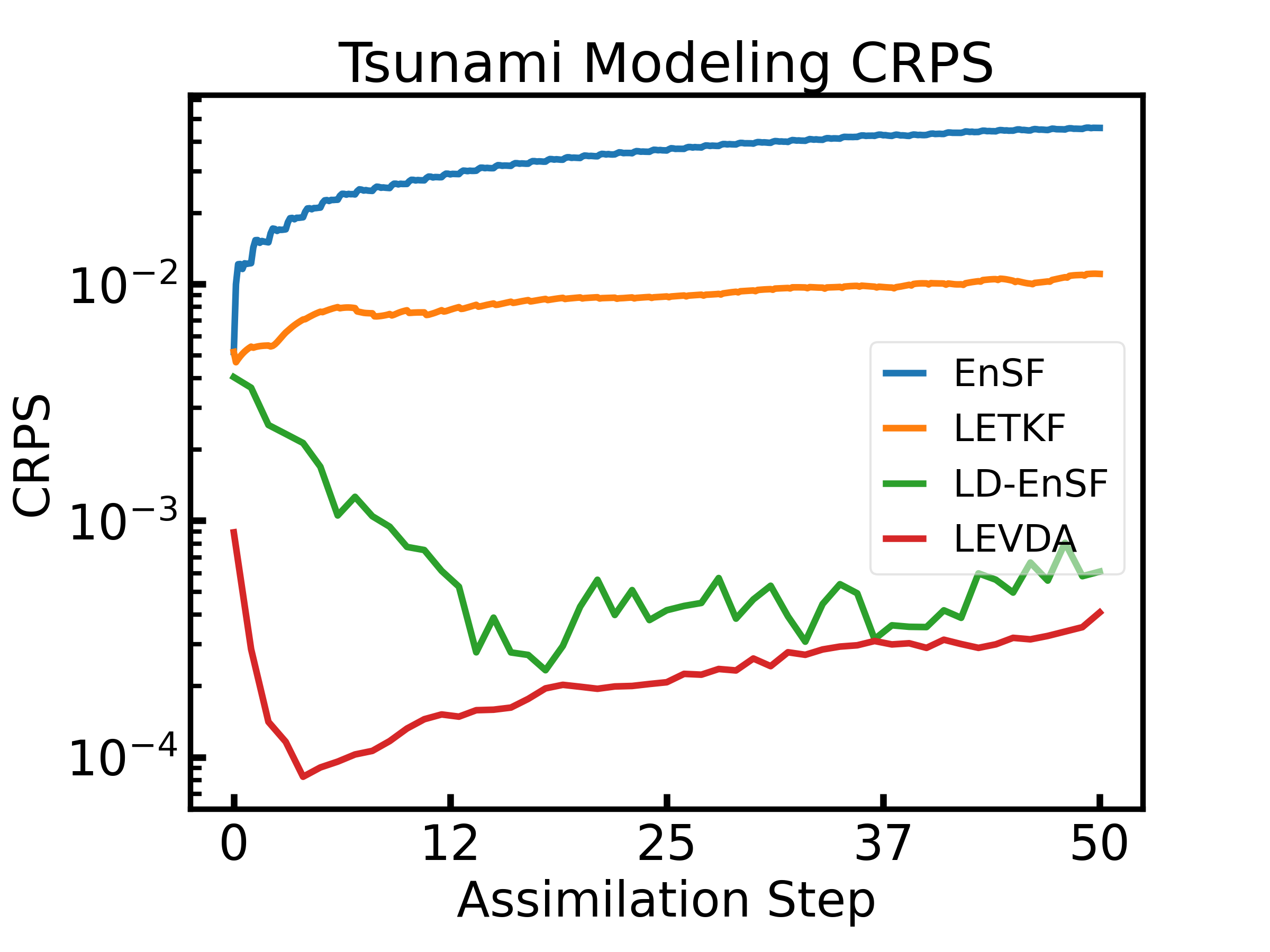}\hspace{-0.2em}
\includegraphics[width=0.33\linewidth]{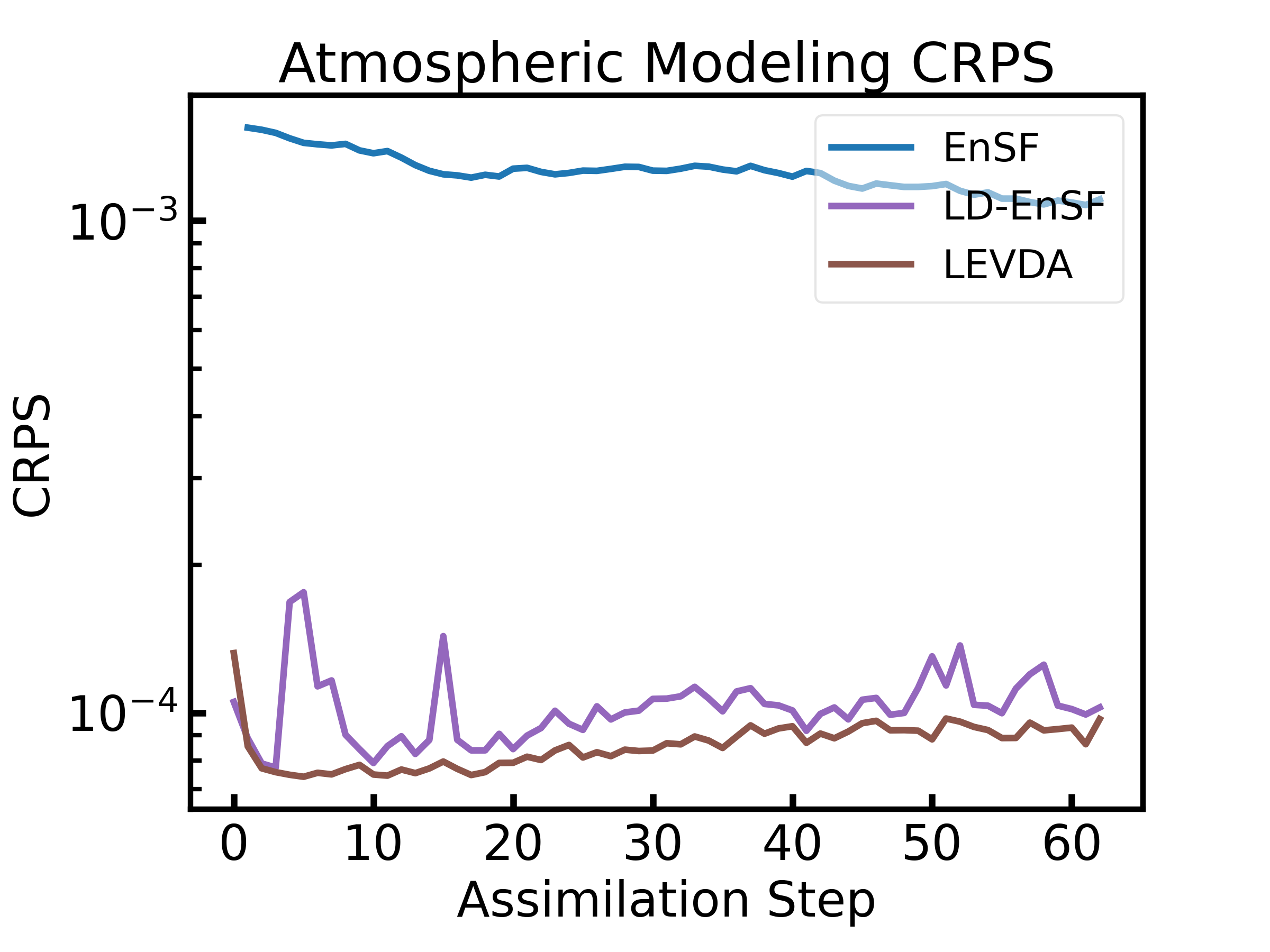}
\caption{Continuous Ranked Probability Score (CRPS) over time for Kolmogorov flow, tsunami modeling, and atmospheric modeling (left-to-right). Lower values represent higher probabilistic accuracy.}
\label{fig:crps_main}
\vspace{-10pt}
\end{figure*}

\paragraph{Continuous Ranked Probability Score.}
Our second uncertainty metric is the Continuous Ranked Probability Score (CRPS), a proper scoring rule that quantifies the discrepancy between the forecast and observed cumulative distribution functions. A lower CRPS indicates superior probabilistic skill, reflecting an optimal balance between predictive accuracy and ensemble dispersion. For high-dimensional states, we compute the CRPS using an energy-score formulation, averaging across spatial dimensions and channels to evaluate performance over time (Figure \ref{fig:crps_main}).

Overall, LEVDA achieves the lowest CRPS and best SER among evaluated baselines, indicating superior probabilistic accuracy. In addition, LEVDA produces SER closer to $1$ than baselines in nearly all settings, reflecting improved calibration rather than merely sharper, under-dispersed posteriors.

\section{Capability Comparison}\label{app:capability}
Table~\ref{tab:capabilities} contrasts LEVDA with existing latent-space ensemble methods and score-based DA approaches. LEVDA distinguishes itself as the only framework in this comparison that simultaneously integrates smoothing, differentiable dynamics, irregular observation windows, and joint parameter estimation.
\begin{table}[H]
\centering
\small
\setlength{\tabcolsep}{3.5pt}
\renewcommand{\arraystretch}{1.15}
\begin{tabular}{lcccccc}
\toprule
\textbf{Capability} 
& \textbf{Latent-EnSF} 
& \textbf{LD-EnSF} 
& \textbf{SDA} 
& \textbf{FlowDAS} 
& \textbf{APPA}
& \textbf{LEVDA}\\
\midrule
Smoothing
& X
& X
& \checkmark
& \checkmark
& \checkmark
& \checkmark \\
\midrule
Differentiable Dynamics
& X
& \checkmark
& X
& X
& X
& \checkmark \\
\midrule
Moving observations 
& X
& X
& \checkmark 
& \checkmark 
& \checkmark
& \checkmark\\
\midrule
Irregular-time observations 
& X
& X
& X
& X
& X
& \checkmark\\
\midrule
Parameter estimation 
& X
& \checkmark 
& X
& X
& X
& \checkmark\\
\bottomrule
\end{tabular}
\caption{Feature comparison across latent and score-based DA methods. Here, ``Differentiable Dynamics'' refers to the use of an explicit learned surrogate for state evolution (as opposed to implicit trajectory priors), while ``Irregular-time Observations'' denotes native support for data at arbitrary time stamps without resampling to a fixed temporal grid.}
\label{tab:capabilities}
\end{table}

\section{Visualization of Experimental Systems}\label{sec:appendix_viz}
\begin{figure}[H]
  \centering
    \includegraphics[width=0.8\linewidth]{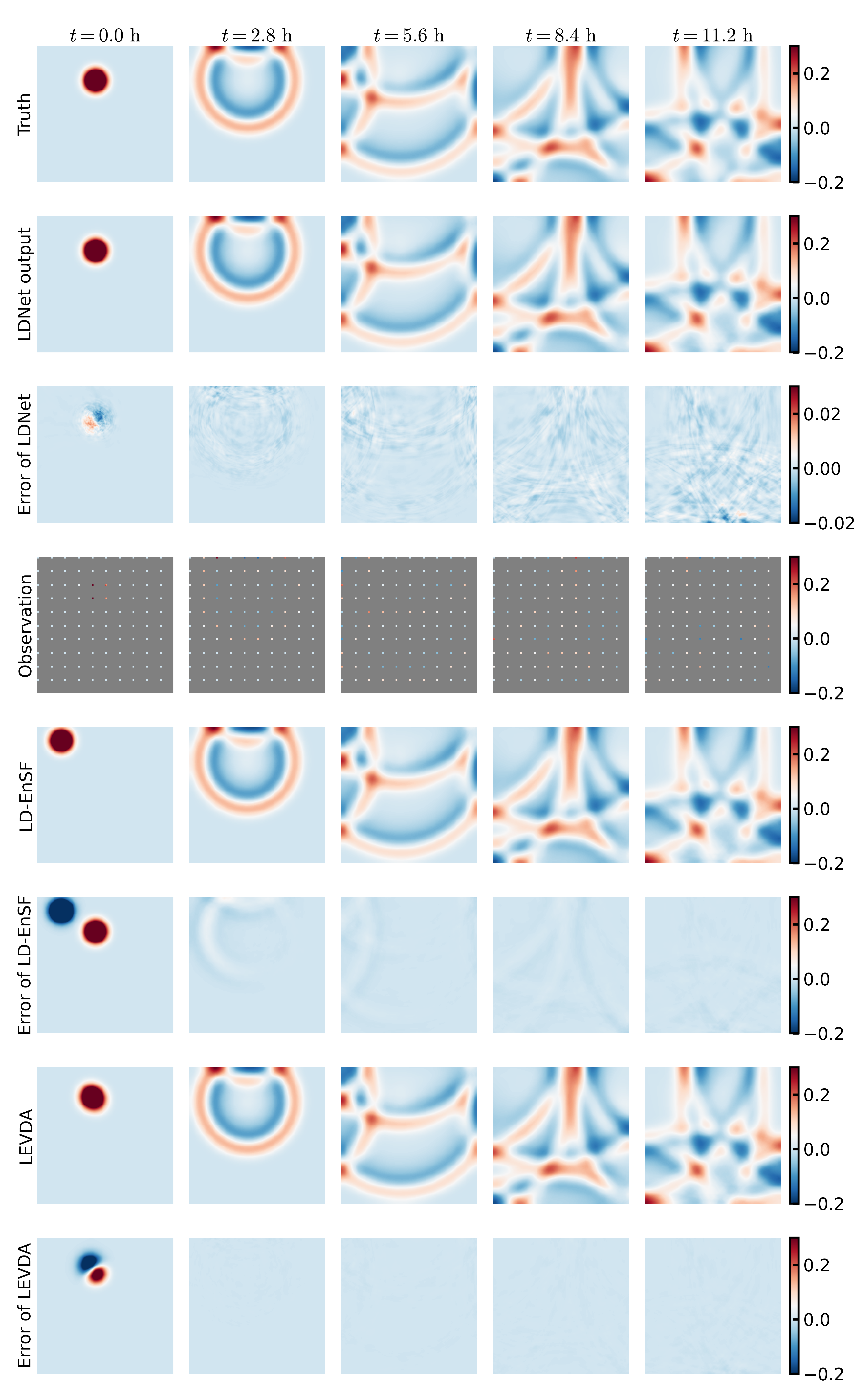}
  \caption{Visualization of surface elevation $\eta$ in tsunami dynamics. The rows display: (1) ground truth; (2) LDNet predictions from a known initial condition; (3) prediction errors; and (4) sparse observations ($10 \times 10$ subsampled from a $150 \times 150$ grid). The remaining rows show assimilation results starting from a deviated initial condition: (5) a single assimilated trajectory from the LD-EnSF ensemble ($N=20$); (6) LD-EnSF error; (7) a single assimilated trajectory from the LEVDA ensemble ($N=20$); and (8) LEVDA error.}
	  \label{fig:sw_visual}
\end{figure}

\begin{figure}[H]
  \centering
    \includegraphics[width=0.8\linewidth]{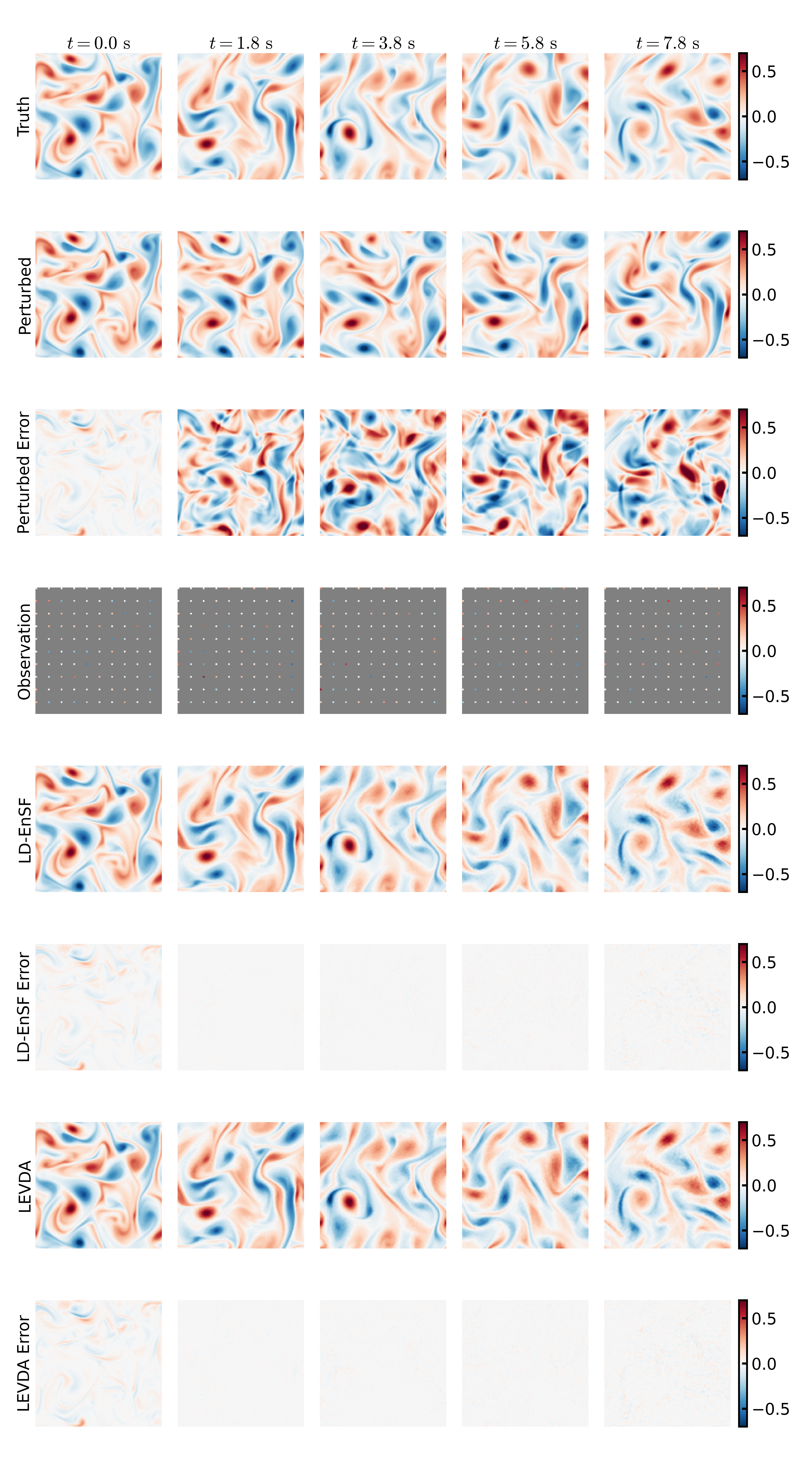}
  \caption{Kolmogorov flow benchmark: vorticity field, comparing ground truth and assimilated reconstruction. The rows display: (1) ground truth; (2) perturbed trajectory with different Reynolds number; (3) error of perturbed trajectory; (4) sparse observations; ($10 \times 10$ subsampled from a $150 \times 150$ grid (5) assimilated results of a single trajectory from the LD-EnSF ensemble ($N=20$); (6) LD-EnSF error; (7) a single assimilated trajectory from the LEVDA ensemble ($N = 20)$; (8) LEVDA error.}
	  \label{fig:kf_visual}
\end{figure}

\end{document}